\documentclass[conference]{IEEEtran}
\usepackage{times}

\usepackage[numbers]{natbib}
\usepackage{multicol}
\usepackage{multirow}
\usepackage[bookmarks=true]{hyperref}
\usepackage{graphicx}
\usepackage{amsmath}
\usepackage{amssymb}
\usepackage{tcolorbox}
\usepackage{xcolor}
\usepackage{cuted}
\usepackage{caption}
\usepackage{subcaption}
\usepackage{booktabs}
\usepackage{stfloats}
\usepackage{listings}
\usepackage{placeins}
\usepackage[hang,flushmargin,bottom]{footmisc}

\newcommand\YAMLcolonstyle{\color{red}\mdseries}
\newcommand\YAMLkeystyle{\color{black}\bfseries}
\newcommand\YAMLvaluestyle{\color{blue}\mdseries}

\lstdefinelanguage{yaml}{
  keywords={true,false,null,y,n},
  keywordstyle=\color{darkgray}\bfseries,
  basicstyle=\YAMLkeystyle\small\ttfamily,
  sensitive=false,
  comment=[l]{\#},
  morecomment=[s]{/*}{*/},
  commentstyle=\color{purple}\ttfamily,
  stringstyle=\YAMLvaluestyle\ttfamily,
  moredelim=[l][\color{orange}]{\&},
  moredelim=[l][\color{magenta}]{\*},
  moredelim=**[il][\YAMLcolonstyle{:}\YAMLvaluestyle]{:},   
  morestring=[b]',
  morestring=[b]"
}

\makeatletter
\renewcommand\footnoterule{%
  \kern 2pt
  \hrule width 0.4\columnwidth
  \kern 6pt
}
\makeatother

\IEEEoverridecommandlockouts

\begin{document}

\title{Semantic-Contact Fields for Category-Level Generalizable Tactile Tool Manipulation}





\author{
\authorblockN{
Kevin Yuchen Ma$^{1,2}$,
Heng Zhang$^{1,3}$,
Weisi Lin$^{3}$,
Mike Zheng Shou$^{*2}$,
and Yan Wu$^{*1}$
\thanks{
Emails: \href{mailto:...}{yuchen\_ma@u.nus.edu}, \href{mailto:...}{HENG018@e.ntu.edu.sg}, \href{mailto:...}{wslin@ntu.edu.sg},
\href{mailto:...}{mikeshou@nus.edu.sg}, \href{mailto:...}{wuy@i2r.a-star.edu.sg}.}
}
\vspace{5pt}

\authorblockA{
$^{1}$Institute for Infocomm Research, A*STAR \quad
$^{2}$Show Lab, National University of Singapore\\
$^{3}$College of Computing and Data Science, Nanyang Technological University
}
\authorblockA{$^{*}$Corresponding authors}


\vspace{-35pt}
}

\maketitle
\begin{strip}
\vspace{-30pt}
    \centering
    \includegraphics[width=1\textwidth]{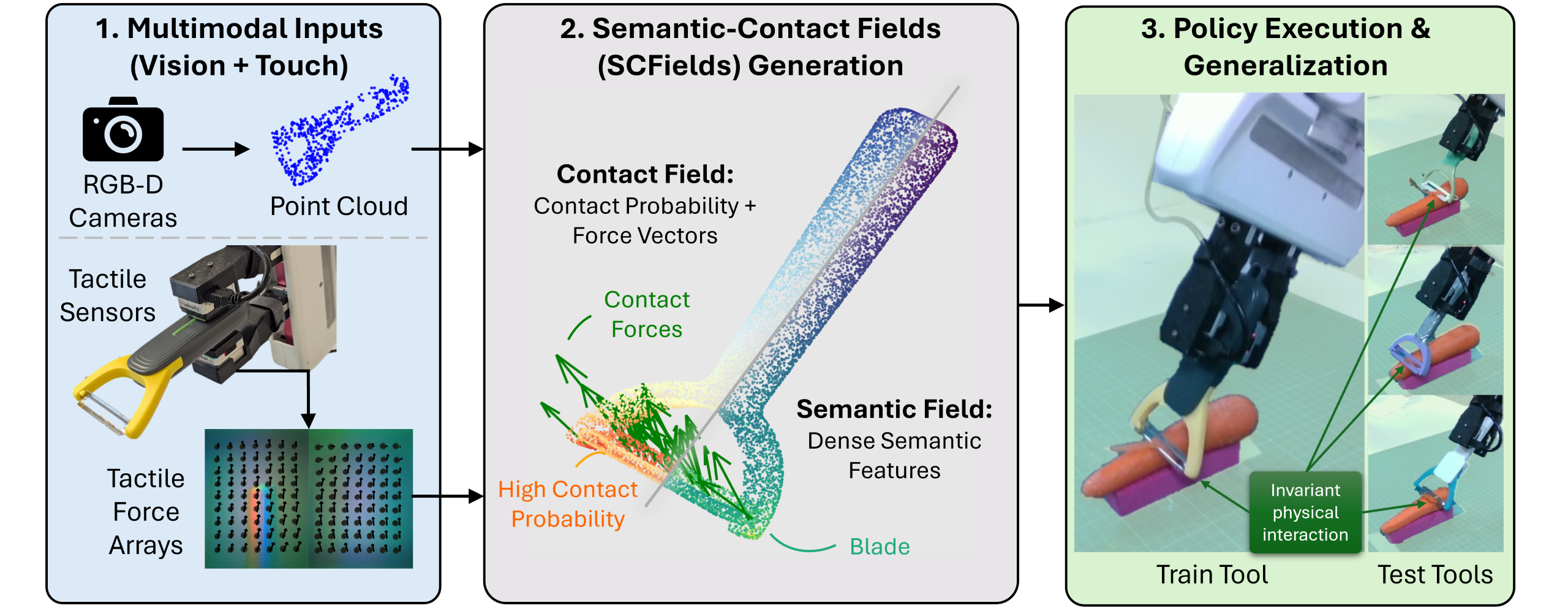}
    \captionof{figure}{\textbf{Semantic-Contact Fields (SCFields) Overview.} 
    \textbf{1. Multimodal Inputs:} The system takes RGB-D observations and tactile readings from GelSight sensors. 
    \textbf{2. SCFields Generation:} Our unified perception module fuses these inputs into a dense point cloud representation containing both category-level semantics (blue/green heatmap) and extrinsic contact force vectors (green arrows).
    \textbf{3. Policy Execution:} A diffusion policy conditioned on the SCFields enables zero-shot generalization to novel tools variants (e.g., peelers of different shapes) in contact-rich tasks by reasoning about functional affordance and contact forces simultaneously.}
    \label{fig:teaser}
    \vspace{-10pt}
\end{strip}

\begin{abstract}

Generalizing tool manipulation requires both semantic planning and precise physical control. Modern generalist robot policies, such as Vision-Language-Action (VLA) models, often lack the physical grounding required for contact-rich tool manipulation. Conversely, existing contact-aware policies that leverage tactile or haptic sensing are typically instance-specific and fail to generalize across diverse tool geometries. Bridging this gap requires learning representations that are both semantically transferable and physically grounded, yet a fundamental barrier remains: diverse real-world tactile data are prohibitive to collect at scale, while direct zero-shot sim-to-real transfer is challenging due to the complex nonlinear deformation of soft tactile sensors.

To address this, we propose \textbf{Semantic-Contact Fields (SCFields)}, a unified 3D representation that fuses visual semantics with dense extrinsic contact estimates, including contact probability and force. SCFields is learned through a two-stage \textbf{Sim-to-Real Contact Learning Pipeline}: we first pre-train on large-scale simulation to learn geometry-aware contact priors, then fine-tune on a small set of real data pseudo-labeled via geometric heuristics and force optimization to align real tactile signals. The resulting force-aware representation serves as the dense observation input to a diffusion policy, enabling physical generalization to unseen tool instances. Experiments on scraping, crayon drawing, and peeling demonstrate robust category-level generalization, significantly outperforming vision-only and raw-tactile baselines. Project page: \href{https://kevinskwk.github.io/SCFields/}{\textcolor{blue}{\texttt{https://kevinskwk.github.io/SCFields}}}.

\end{abstract}    

\IEEEpeerreviewmaketitle

\section{Introduction}
\label{sec:intro}

Tool use represents a hallmark of intelligence, extending a robot's physical capabilities beyond its own embodiment. However, achieving robust category-level generalization in tool manipulation remains challenging. Effectively manipulating diverse tool variants requires a dual understanding: a semantic grasp of \textit{where} to hold and apply the tool (functional affordance) and a physical mastery of \textit{how} to regulate interaction forces (contact dynamics). 

While recent advances in large-scale robotic learning have produced generalist policies capable of interpreting high-level semantic commands, these vision-centric models remain physically naive. Methods relying solely on visual or 3D semantic representations, such as GenDP \cite{wang2024gendp}, can generalize geometrically across tool variants, but fail in contact-rich tasks that are visually ambiguous and require precise force regulation. Conversely, policies that leverage tactile or haptic sensing \cite{bi2025vlatouch, liu2025forcemimic} are adept at managing local contact, but are typically instance-specific. Because they map tactile signals directly to actions without an intermediate generalized representation, they fail to adapt when the tool's geometry changes.

Our approach is grounded in a key physical insight: while the global geometry of tools within a category varies significantly, the physical interaction at the "effective part"—such as the blade of a peeler—remains invariant. Therefore, an explicit representation of \textit{extrinsic contact}---the contact between the tool and the environment---can provide a geometry-aware physical abstraction for transferring manipulation skills across diverse tool instances. A promising candidate for this is the contact field \cite{higuera2023ncf}, which maps tactile feedback onto the tool's surface. However, for contact-rich tool use, contact localization alone is insufficient: the policy must also infer the magnitude and direction of interaction forces needed for regulation. Learning such representations also presents a dilemma: collecting diverse real-world tactile data to cover all geometries is prohibitively expensive, yet training entirely in simulation introduces a severe Sim-to-Real gap. Simulating the non-linear deformation of soft tactile sensors is notoriously difficult, leading models trained solely in simulation to hallucinate phantom forces or miss contacts entirely when deployed.

To bridge this gap, we propose \textbf{Semantic-Contact Fields (SCFields)} (Figure \ref{fig:teaser}), a unified force-aware 3D representation trained via a two-stage Sim-to-Real framework. We decompose the contact estimation problem into learning \textit{general geometry and interaction physics} (invariant contact distributions) and \textit{real-world sensory alignment} (sensor-specific signal interpretation). Accordingly, our pipeline operates in two distinct stages. First, we leverage large-scale simulation across diverse tool geometries to learn geometry-aware contact priors for the invariant contact physics. Second, to address the reality gap, we introduce a \textbf{Real-World Alignment} stage. We generate pseudo-labels using geometric heuristics and analytical methods from a small set of simple real-world interactions. This alignment step adapts the simulation-trained model to interpret real sensor responses while preserving the generalizable physics learned in simulation. Importantly, the data used for this alignment stage is collected concurrently with the demonstrations used for imitation learning policy training, meaning no additional, separate data collection effort is required.

Our specific contributions are as follows:
\begin{itemize}
    \item We propose \textbf{Semantic-Contact Fields (SCFields)}, an invariant 3D representation that fuses semantic features from pre-trained vision models with dense extrinsic contact probability and force estimates.
    \item We introduce a two-stage \textbf{Sim-to-Real training pipeline} that combines the scalability of simulation with the fidelity of real-world data. By pre-training on diverse simulated tools and fine-tuning with heuristic-labeled real data, we achieve robust dense contact estimates generalizable to unseen tool variants.
    \item We evaluate our approach on a Franka Panda robot with Gelsight Mini tactile sensors across scraping, crayon drawing, and peeling, demonstrating category-level generalization to unseen tool instances and environmental variations.
\end{itemize}

\section{Related Work}

\begin{figure*} [ht]
    \centering
    \vspace{-10pt}
    \includegraphics[width=\textwidth]{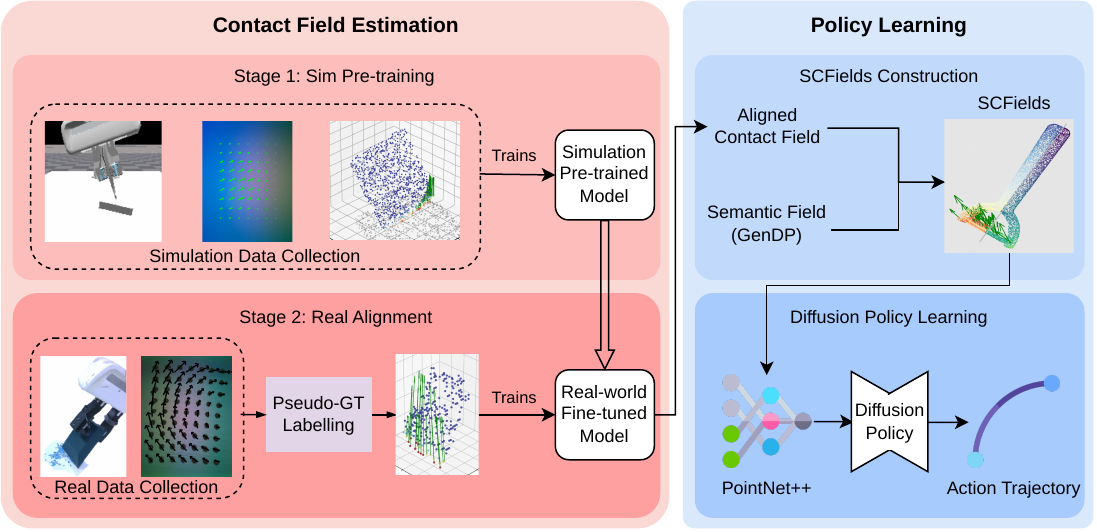}
    \caption{\textbf{Method Overview.} \textbf{Left}: Contact Field Learning (\ref{subsec:contact_field_estimation}) Stage 1 learns general geometry-aware contact priors from simulated data; Stage 2 aligns the model to real tactile sensor responses using pseudo-labeled real data. \textbf{Right}: Policy Learning (\ref{subsec:scfield_policy}) A Diffusion Policy is conditioned on the resulting SCFields to achieve robust tool manipulation.}
    \label{fig:overview}
    \vspace{-15pt}
\end{figure*}

\subsection{Generalizable 3D Manipulation and Tool Use}

A key limitation of 2D image-based policies is their sensitivity to viewpoint changes. This has led to a surge in 3D-centric policies that operate directly on point clouds or neural fields \cite{ze20243ddiffusionpolicy, gervet2023act3d, ke20243ddiffuseractor, zhen20243dvla}, providing improved spatial reasoning. To enable category-level generalization, recent works have adopted semantic-centric representations. Methods such as D3Field \cite{wang2023d3fields}, GenDP \cite{wang2024gendp}, and S$^2$-Diffusion \cite{yang2025s2diffusion} use spatial or 3D semantic representations to improve category-level generalization. While these methods achieve strong geometric generalization, they remain purely vision-based and physically naive, struggling in contact-rich tasks where visual semantics alone are insufficient.

In tool manipulation, generalization requires understanding both tool geometry and functional interaction. Prior works address this by learning structured correspondences between tool instances. Some approaches focus on dense alignment, predicting motion fields or 6D functional poses to transfer manipulation motions across tools \cite{seita2023toolflownet, wang2024tooleenet}. Other methods utilize sparser representations to enable one-shot skill transfer, modeling tools via functional keypoints \cite{tang2025functo, tang2025mimicfunc} or leveraging vision-language models to identify affordance regions \cite{singh2025afford2act, liu2024moka}. 
While effective at identifying \textit{where} the functional part of a tool is, these methods often do not model \textit{how} contact should be regulated during execution, such as the force direction and magnitude required to peel or scrape.

Attempts to bridge this gap by combining vision and touch have yielded promising but limited results.
3D-ViTac \cite{huang20243dvitac} integrates tactile readings as occupancy points into a scene point cloud, providing a useful geometric fusion mechanism, but not an explicit estimate of dynamic extrinsic force on the tool surface.
Other works rely on wrist force-torque sensors for compliance \cite{liu2025forcemimic, he2024foarforceawarereactivepolicy, liu2025factr}. 
However, for tool manipulation, wrist-mounted sensing provides only a net wrench at the robot wrist; inferring the critical distributed contact location and force at the tool tip remains ambiguous because of tool leverage, grasp variation, and external contacts.
Our work addresses this gap by representing extrinsic contact directly on the tool point cloud, combining semantic invariance with per-point contact probability and force estimates.

\subsection{Tactile Perception and Sim-to-Real Transfer}

The primary challenge in leveraging the tactile modality is converting high-dimensional, noisy sensor data into a useful representation. Vision-based sensors like GelSight \cite{yuan2017gelsight} provide rich, high-resolution topography, while taxel-based sensors offer direct force maps. Recent advances in tactile representation learning, such as T3 \cite{zhao2024transferable} and Sparsh \cite{higuera2024sparsh}, utilize self-supervision to learn robust encoders. 
However, these representations typically encode intrinsic tactile observations at the sensor surface, rather than grounding contact on the external tool geometry. This limits their direct use for tool manipulation, where the policy must reason about where and how the tool contacts the environment.

Tactile manipulation is further hindered by the significant gap between simulated and real tactile physics. General tactile simulators \cite{wang2022tacto, si2022taxim, akinola2025tacsl} have made strides in modeling sensor deformation and optical properties. However, accurately capturing fine-grained contact mechanics—such as friction coefficients, hysteresis, and soft-body dynamics—remains computationally intensive and difficult to calibrate. As a result, policies trained purely in simulation \cite{ding2021sim, lin2023bi, akinola2025tacsl} often struggle to generalize to the unmodeled physical variations of the real world. To address this, we use simulation to learn geometry-aware contact priors, then fine-tune the contact estimator on a small set of pseudo-labeled real data to align real tactile sensor responses without requiring high-fidelity tactile simulation.

\subsection{Extrinsic Contact Estimation}

To manipulate tools effectively, a robot must understand \textit{extrinsic contact}—the interaction between the tool and the environment. Prior work has attempted to estimate this property through various means. Vision-only approaches \cite{kim2023im2contact} can infer likely contact regions but remain ambiguous without tactile feedback, while analytical methods \cite{ma2021extrinsic} often rely on known object geometries and specific exploratory motions.

The most promising recent direction involves learning implicit contact representations. Neural Contact Fields (NCF) \cite{higuera2023ncf, higuera2023ncfpolicy} learn a continuous function mapping surface coordinates to contact probabilities. However, NCF assumes a fixed in-hand pose of the tool relative to the sensor, which breaks down during dynamic manipulation where grasp adaptation occurs. More recent work like VitaScope \cite{vitascope2025} relaxes this assumption by jointly estimating the tool in-hand pose and the extrinsic contact. However, VitaScope requires a known tool mesh, making category-level generalization to previously unseen tool geometries difficult. Furthermore, these methods typically model contact probability via geometric proximity, ignoring the magnitude of contact forces. In contrast, SCFields represents extrinsic contact directly on the observed tool point cloud and predicts both contact probability and force, enabling force-aware manipulation of novel tool instances without assuming a fixed in-hand pose or known mesh.
\section{Methods}

Our goal is to learn a manipulation policy that generalizes to unseen tool variants by leveraging contact-rich dynamics. To achieve this, we propose a two-stage method (Figure \ref{fig:overview}):
\begin{enumerate}
    \item \textbf{Contact Field Learning:} We train a multimodal perception model $f_{\phi}$ to estimate a dense Extrinsic Contact Field on the tool surface. This model is pre-trained in simulation to learn geometry-aware contact priors and fine-tuned on real-world data for domain alignment.
    \item \textbf{Policy Learning:} We construct a unified state representation by fusing these estimated contact probabilities and forces with 3D Semantic Fields \cite{wang2024gendp}. This fused representation conditions a diffusion policy $\pi_{\theta}$ capable of zero-shot transfer to novel tool instances.
\end{enumerate}

\subsection{Problem Formulation}
\label{subsec:problem_formulation}

We formulate the system as two distinct learning problems:

\textbf{1. Extrinsic Contact Field Estimation:} We learn a perception mapping $f_{\phi}$ that transforms raw observations—tool and environment point clouds ($P_{\text{tool}}, P_{\text{env}}$), tactile sensor readings ($T$), and proprioceptive state ($\boldsymbol{q}$)—into a dense extrinsic contact field $F_c$ over the tool surface. This field assigns a contact probability $c_i$ and a 3D force vector $\boldsymbol{f}_i$ to every point $p_i$ in the tool point cloud $P_{\text{tool}}$. We predict both quantities because they serve different roles: $c_i$ localizes the plausible support of tool-environment contact, while $\boldsymbol{f}_i$ encodes the local interaction direction and magnitude needed for force regulation.

\textbf{2. Generalizable Policy Learning:} We treat manipulation as conditional generation. We learn a policy $\pi_{\theta}(a_{t:t+H} | O_t)$ that predicts a sequence of actions $a$ given observation $O_t$. The core challenge is designing an $O_t$ that is invariant to instance-specific geometry while retaining high-fidelity physical feedback.

\subsection{Contact Field Estimation}
\label{subsec:contact_field_estimation}

\subsubsection{Model Architecture}
\label{subsubsec:model_architecture}

\begin{figure}
    \centering
    \includegraphics[width=1\linewidth]{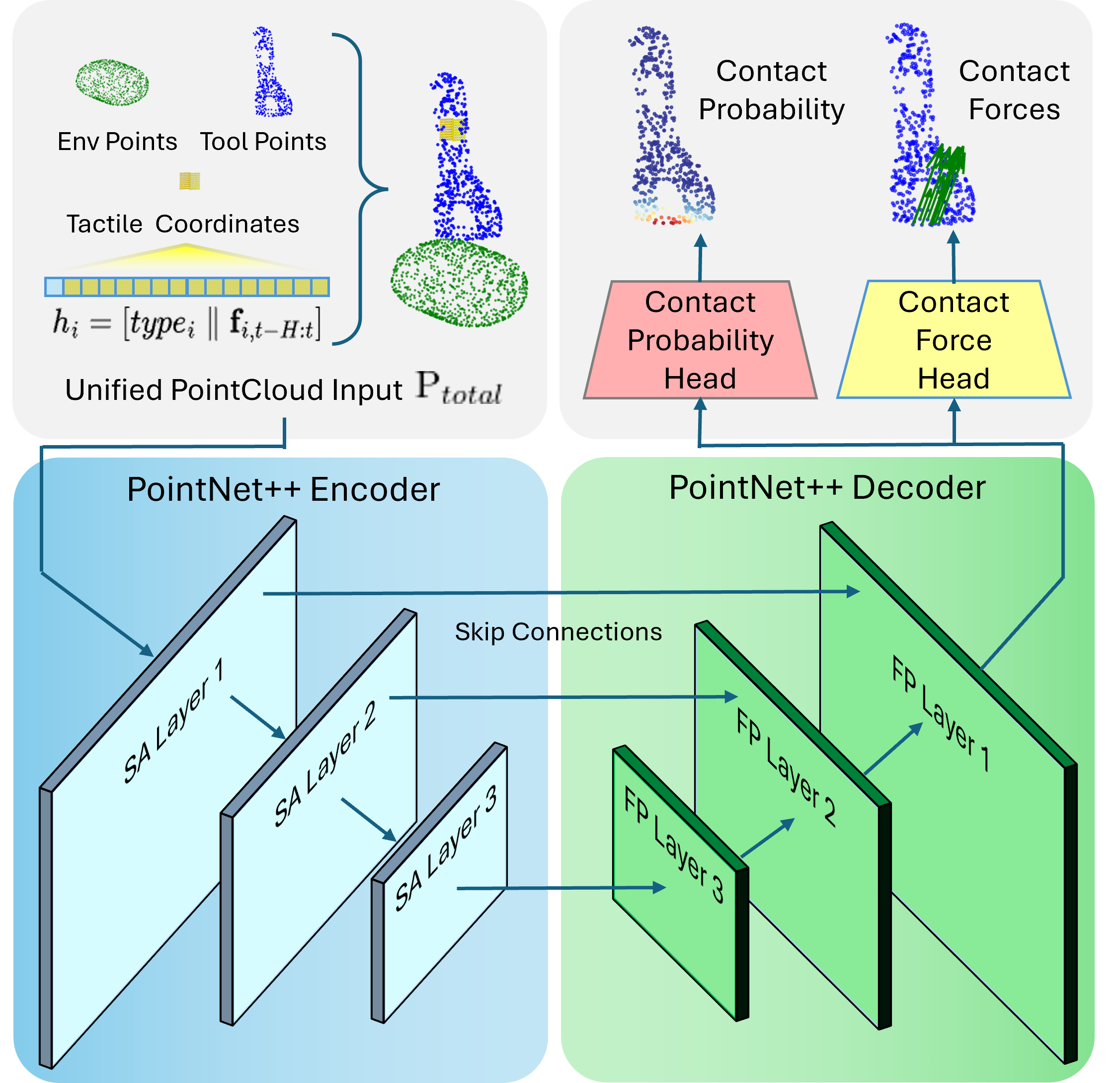}
    \caption{
    Contact field model architecture. The network fuses tactile markers and force arrays with dense object geometry in a unified point cloud input to predict contact fields.}
    \label{fig:network_architecture}
    \vspace{-10pt}
\end{figure}

Unlike prior works that process vision and touch with separate encoders \cite{higuera2023ncf}, we employ a unified \textbf{Tactile-as-PointCloud} architecture similar to 3D-ViTac \cite{huang20243dvitac}. We treat tactile signals as 3D geometric entities, fusing them directly into the scene geometry using a PointNet++ \cite{qi2017pointnet++} framework (Figure \ref{fig:network_architecture}).

\noindent\textbf{Unified Input Representation.} The input is a composite point cloud $P_{total} = P_{obj} \cup P_{env} \cup P_{tactile}$, where $P_{obj}$ and $P_{env}$ represent the sampled tool and environment surfaces, and $P_{tactile}$ represents the 3D coordinates of the tactile sensor markers projected into the world frame.

Each point $p_i \in P_{total}$ is augmented with a feature vector $h_i = [ type_i \parallel \mathbf{f}_{i, t-H:t} ]$. Here, $type_i$ encodes the source (Object, Env, Tactile), and $\mathbf{f}_{i, t-H:t}$ encodes a $H$-step history of marker displacements. This representation allows the network to implicitly learn the relationship between tool geometry, sensor deformation, and contact location without a dedicated pose encoder.

\noindent\textbf{Network Architecture.}
We process $P_{total}$ using a standard PointNet++ encoder-decoder. The encoder fuses sparse tactile signals with dense tool geometry based on spatial proximity. The decoder upsamples the features back to the resolution of $P_{obj}$, effectively propagating localized sensor information to the entire tool surface. Two parallel heads then predict the scalar contact probability $c_i \in [0, 1]$ and regress the extrinsic force vector $\mathbf{f}^{ext}_i \in \mathbb{R}^3$ for each point. Detailed layer configurations are provided in Appendix \ref{app:architecture}.

\vspace{3pt}

\subsubsection{Simulation Data Generation}
\label{subsubsec:sim_data_generation}

Training requires dense ground-truth (GT) labels for contact locations and forces, which are physically inaccessible in the real world. We address this by generating a large-scale synthetic dataset (300 tool instances, 320,000 frames) using a multi-simulator pipeline, inspired by \cite{vitascope2025}.

\noindent\textbf{Simulation Pipeline.}
We construct a simulation environment comprising a Franka Emika Panda robot and diverse procedurally generated tools (scrapers, crayons, peelers). Data collection proceeds in two steps:
\begin{enumerate}
    \item \textbf{Interaction (IsaacGym + TacSL):} We use IsaacGym \cite{makoviychuk2021isaac} for high-throughput rigid-body dynamics and TacSL \cite{akinola2025tacsl} to simulate the specific force field and depth data of GelSight sensors.
    \item \textbf{Labeling (Open3D + PyBullet):} Since accurate contact locations and forces are not directly accessible in IsaacGym, we employ a replay strategy. We replicate the scene in Open3D to compute signed distance functions (SDF) for soft contact probability, and replay interactions in PyBullet to extract discrete contact forces, which are then extrapolated to the dense point cloud.
\end{enumerate}
This pipeline produces dense contact labels for 300 unique tool geometries, providing broad geometric and contact variation for pre-training. Further details on the replay and extrapolation logic are in Appendix \ref{app:sim_data_labeling}.

\vspace{3pt}

\subsubsection{Real-World Pseudo-GT Generation}
\label{subsubsec:real_pseudo_gt}

Despite filtering and calibration, a significant gap remains between simulated tactile fields and real marker displacements. Furthermore, obtaining dense ground-truth extrinsic contact fields directly in the real world is extremely difficult, as it would require instrumenting the entire surface of arbitrary tools with high-resolution force sensors. To bridge these challenges, we introduce a \textbf{Real-World Alignment} stage. We collect a small real-world alignment dataset and generate pseudo-ground-truth labels using geometric heuristics and analytical force optimization.

\noindent\textbf{Heuristic Contact Probability.}
We constrain data collection to a structured task: scraping a flat surface with known table height $z_{table}$. We first identify geometrically plausible contact candidates $C_{candidate} \subset P_{obj}$ using a height threshold ($p_z < z_{table} + \epsilon$). To eliminate false positives (e.g., the tool hovering near the surface without touching), we apply a signal-based gating filter. We compute the mean magnitude of tactile marker displacements relative to the initial undeformed frame. A frame is labeled as "in contact" only if this mean delta signal exceeds a calibrated noise threshold. For these valid frames, points in $C_{candidate}$ are assigned soft contact probability labels $c_i \in [0, 1]$ inversely proportional to their distance from the table surface.

\noindent\textbf{Analytical Contact Force Optimization.}
To estimate dense force vectors $\mathbf{f}^{ext}_i$ without ground-truth from external force sensors, we solve a convex optimization problem that explains the observed tactile net wrench $\mathbf{W}_{tac}$ using a distribution of point forces $\mathbf{f}$ at candidate contact points. We formulate this as a Second-Order Cone Program (SOCP):

\vspace{-15pt}

\begin{align}
    \min_{\mathbf{f}} \quad & \left\| \mathbf{G} \mathbf{f} - \mathbf{W}_{tac} \right\|_2^2 + \lambda \sum_{i \in C_{candidate}} \frac{\| \mathbf{f}_i \|_2^2}{c_i + \epsilon} \\
    \text{s.t.} \quad & \| \mathbf{f}_i \|_2 \leq 2 (\mathbf{f}_i \cdot \mathbf{n}_i) \quad \forall i
\end{align}

where $\mathbf{G}$ is the grasp matrix mapping point forces to the gripper frame. The objective minimizes wrench discrepancy while regularizing force magnitudes inversely to their heuristic contact probability $c_i$, favoring geometrically likely contact points. The constraint enforces physical plausibility by bounding force magnitude by twice its projection onto the inward normal $\mathbf{n}_i$. This ensures forces are compressive and lie within a $\sim 60^{\circ}$ friction cone, preventing unrealistic "pulling" or shear. The problem is solved efficiently using the ECOS solver \cite{Domahidi2013ecos}.

\vspace{3pt}

\subsubsection{Two-Stage Model Training}
\label{subsubsec:two_stage_training}

We employ a two-stage strategy to transfer physical priors from simulation to the real world. Both stages minimize a composite loss $\mathcal{L}_{total} = \lambda_{prob} \mathcal{L}_{prob} + \lambda_{force} \mathcal{L}_{force}$. We use Focal Loss \cite{lin2017focalloss} for $\mathcal{L}_{prob}$ to handle class imbalance, and a combination of Adaptive Weighted MSE (magnitude) and Cosine Similarity (direction) for $\mathcal{L}_{force}$. Detailed training and loss function hyperparameters are presented in Appendix \ref{app:training_params}.

\noindent\textbf{Stage 1: Sim Pre-training.} The model is first trained on the large-scale simulation dataset with extensive domain randomization. This establishes the fundamental mapping between tool geometry and force distribution across a wide range of tool variants and interaction poses.

\noindent\textbf{Stage 2: Real-World Alignment.} We fine-tune the model on the pseudo-labeled real-world dataset. Since this real-world set is small and collected under constrained conditions, we apply random translation and rotation augmentations to the input point clouds during training to enhance data diversity, preventing overfitting to the specific collection poses and ensuring that the learned sensor alignment generalizes robustly to varied spatial configurations. We use a reduced learning rate to adapt to the real sensor characteristics while preserving the general geometry-aware contact priors learned in simulation.

\subsection{Semantic-Contact Fields for Policy Learning}
\label{subsec:scfield_policy}

We construct \textbf{Semantic-Contact Fields (SCFields)} as a unified 3D observation that fuses semantic information with aligned extrinsic contact estimates.
The policy observation $s_t$ is a dense feature field over the object point cloud $P_{obj}$. Each point $p_i$ carries a feature vector $x_i = [ \mathbf{f}^{ext}_i \parallel c_i \parallel S_i ]$, where:
\begin{itemize}
    \item $\mathbf{f}^{ext}_i, c_i$: \textbf{Contact Field} with contact force and probability estimates from our fine-tuned contact estimator.
    \item $S_i$: \textbf{3D Semantic Fields} adapted from \cite{wang2024gendp} extracted from a pre-trained vision backbone, capturing functional semantics such as "blade" and "handle", providing geometric invariance.
\end{itemize}

We implement the manipulation policy $\pi_{\theta}$ using a 3D Diffusion Policy framework based on \cite{wang2024gendp}. The dense SCFields point cloud $\{p_i, x_i\}$ is first processed by a PointNet++ backbone to extract a global feature vector that aggregates both semantic and physical information. This feature vector is then fed into the diffusion policy's denoising network as the conditioning input. During inference, the policy iteratively denoises Gaussian noise into a sequence of end-effector actions $a_{t:t+H}$, conditioned on both functional semantics and force-aware contact estimates from SCFields.
\begin{table*}[ht]
    \centering
    \begin{minipage}[t]{0.41\textwidth}
        \centering
        \caption{Sim Evaluation: Architecture Capacity}
        \label{tab:sim_metrics}
        \small
        \resizebox{\linewidth}{!}{
        \begin{tabular}{lcc}
            \toprule
            Model & F1 Score$\uparrow$ & Force MSE$\downarrow$ \\
            \midrule
            NCF \cite{higuera2023ncf} & 0.043 & N/A \\
            No-Tactile & 0.539 & 0.0146 \\
            Ablation - 2D Tactile Enc. & 0.531 & 0.0147 \\
            Ablation - BCE Loss & 0.123 & \textbf{0.0146} \\
            Ablation - No Cont. Prob. & N/A & 0.0158 \\
            \textbf{Ours (Sim-Only)} & \textbf{0.587} & 0.0147 \\
            \bottomrule
        \end{tabular}
        }
    \end{minipage}\hfill
    \begin{minipage}[t]{0.59\textwidth}
        \centering
        \caption{Real-World Evaluation: Alignment \& Generalization}
        \label{tab:real_alignment}
        \small
        \resizebox{\linewidth}{!}{
        \begin{tabular}{lcc|cc}
            \toprule
            & \multicolumn{2}{c|}{\textbf{Scrapers (Seen Class)}} & \multicolumn{2}{c}{\textbf{Crayons (Unseen in Real)}} \\        
            Model & F1 Score$\uparrow$ & Force MSE$\downarrow$ & F1 Score$\uparrow$ & Force MSE$\downarrow$ \\ 
            \midrule
            Sim-Only & 0.002 & 0.0435 & 0.008 & 0.0284 \\
            Real-Only & 0.458 & \textbf{0.0221} & 0.614 & 0.0106 \\
            No-Tactile & 0.411 & 0.0432 & 0.552 & 0.0115 \\
            No Cont. Prob. & N/A & 0.0377 & N/A & 0.0089 \\
            \textbf{Ours (Aligned)} & \textbf{0.534} & 0.0254 & \textbf{0.657} & \textbf{0.0085} \\ 
            \bottomrule
        \end{tabular}
        }
    \end{minipage}
\end{table*}

\begin{figure*} [h]
    \centering
    \includegraphics[width=\linewidth]{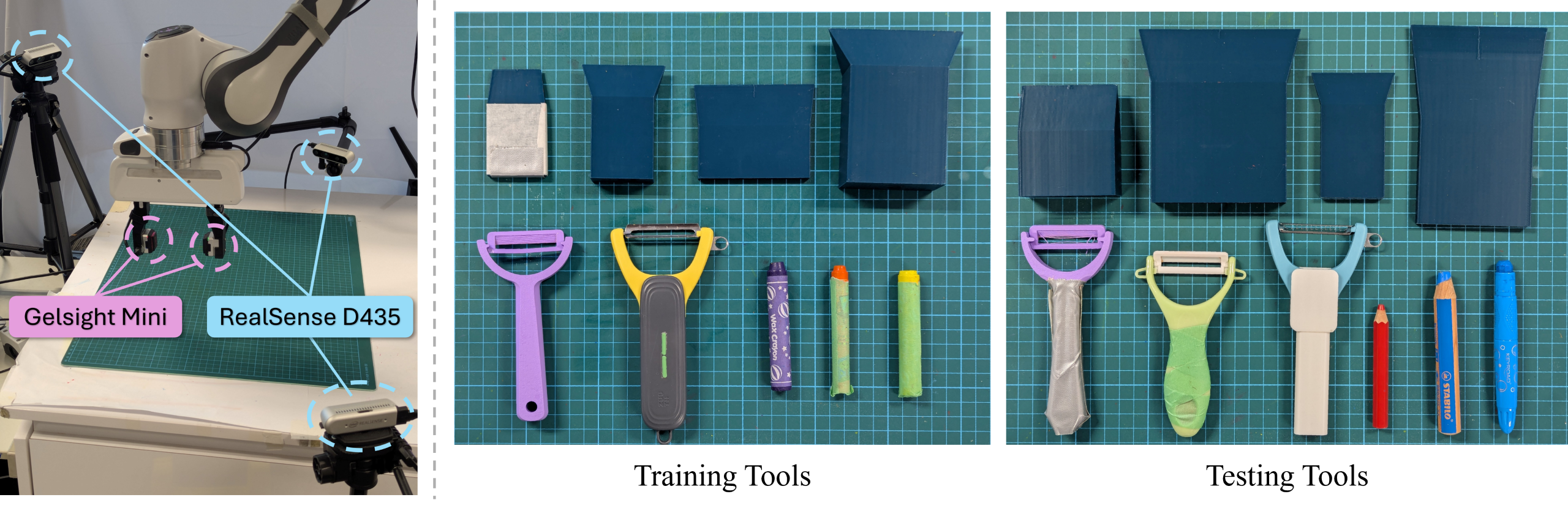}
    \caption{\textbf{Left}: Real robot experiment setup: We use a Franka Emika Panda robot with 2 GelSight Mini tactile sensors mounted on the gripper fingers, and 3 RealSense D435 cameras to capture RGB-D observations. \textbf{Right}: Training and Testing Tools}   
    \label{fig:setup}
    \vspace{-10pt}
\end{figure*}

\section{Experiments}
\label{sec:experiments}

We design our experiments to evaluate two main hypotheses: (1) Does our unified \textit{Tactile-as-PointCloud} architecture, combined with real-world alignment, produce accurate contact estimates that generalize to unseen tools? (2) Does the resulting \textbf{Semantic-Contact Fields (SCFields)} enable a diffusion policy to perform contact-rich manipulation tasks that are robust to environmental variations and novel tool instances?

\subsection{Contact Field Model Evaluation}
\label{subsec:exp_contact_field}

We first validate the contact field perception module ($f_{\phi}$) in isolation to ensure it provides reliable contact probability and force vector estimates before integrating it into the policy loop. 
For contact-field metrics, we report aggregate performance in the main text and provide confidence intervals and significance tests in Appendix~\ref{app:statistics}.

\noindent\textbf{Models Compared.} To isolate the contributions of our architecture and training pipeline, we categorize comparisons as follows:

\noindent\textit{1. Sim-to-Real Training Strategies (Validating the Pipeline):}
\begin{itemize}
    \item \textbf{Ours (Aligned):} Pre-trained on large-scale simulation, then fine-tuned on the small real-world scraper dataset.
    \item \textbf{Sim-Only:} Trained exclusively on simulation data. Evaluates zero-shot transfer and the magnitude of the sim-to-real gap.
    \item \textbf{Real-Only:} Trained from scratch on the small real-world scraper dataset. Tests if physical priors from simulation are necessary given limited real data.
\end{itemize}

\noindent\textit{2. Baselines \& Ablations (Validating the Architecture):}
\begin{itemize}
    \item \textbf{No-Tactile:} Identical architecture with tactile marker features masked out. This baseline relies only on observed tool/environment geometric ($P_{obj}$, $P_{env}$), testing the necessity of dynamic tactile feedback.
    \item \textbf{NCF (Neural Contact Fields)~\cite{higuera2023ncf}:} A baseline implicit representation that predicts contact probabilities but lacks explicit force vector regression.
    \item \textbf{Ablation - 2D Tactile Encoder:} Replaces our point-cloud fusion with a standard CNN encoder for the tactile force arrays, concatenated with the global point cloud feature. This tests the benefit of our \textit{Tactile-as-PointCloud} fusion strategy.
    \item \textbf{Ablation - Loss Function:} Replaces Focal Loss with standard BCE Loss to evaluate robustness to class imbalance.
\end{itemize}

\subsubsection{Evaluation 1: Architecture Validation (Sim-to-Sim)}
We first verify the architecture's capacity to learn complex contact physics using a held-out test set from our simulation dataset. This controlled setting allows us to compare architectural choices without domain shift noise. We report the \textbf{F1 Score} for binary contact detection and the \textbf{Mean Squared Error (MSE)} for force vector regression.

As shown in Table~\ref{tab:sim_metrics}, our \textbf{Tactile-as-PointCloud} architecture trained with Focal Loss achieves the best contact F1, supporting the importance of Focal Loss and preserving the 3D spatial structure of tactile markers under severe contact-class imbalance. NCF performs poorly, likely because its fixed-pose formulation is sensitive to varying in-hand tool poses. The Force MSE values are similar for several models in simulation, suggesting that force regression in the synthetic setting is less discriminative due to limited simulated tactile/contact fidelity. However, removing the contact probability head increases Force MSE, indicating that contact probability provides a useful spatial support signal for force regression.

\vspace{3pt}

\subsubsection{Evaluation 2: Real-World Alignment Accuracy}
We assess the efficacy of our pipeline on real-world data. We evaluate on a held-out set of \textbf{Scrapers} (Seen in Alignment) and a set of \textbf{Crayons} (Unseen in Alignment, but Seen in Sim). Ground truth is generated via our heuristic labeling pipeline.

Table~\ref{tab:real_alignment} highlights the severity of the sim-to-real gap: the \textbf{Sim-Only} model fails almost completely on real tactile inputs. The \textbf{Real-Only} model fits the seen scraper alignment data but generalizes less reliably to crayons, indicating that limited real data alone does not provide sufficient geometric contact priors. The \textbf{No-Tactile} baseline detects some contact from geometry but has substantially worse force prediction, showing that geometric proximity is insufficient for loaded contact estimation. In contrast, \textbf{Ours (Aligned)} transfers the sensor alignment learned from scraper interactions to crayons, supporting the separation between simulation-learned contact priors and real tactile-signal alignment.
The \textbf{No Contact Probability} ablation further clarifies the role of the probability head. Without contact probability, force estimation degrades substantially on real scrapers, where contact is distributed along an edge, but changes little on crayons, where contact is closer to point-like. This suggests that contact probability is most useful as a spatial support estimator for extended or ambiguous contact regions, rather than as a replacement for force prediction.

\begin{figure*}[h]
    \centering
    \includegraphics[width=\linewidth]{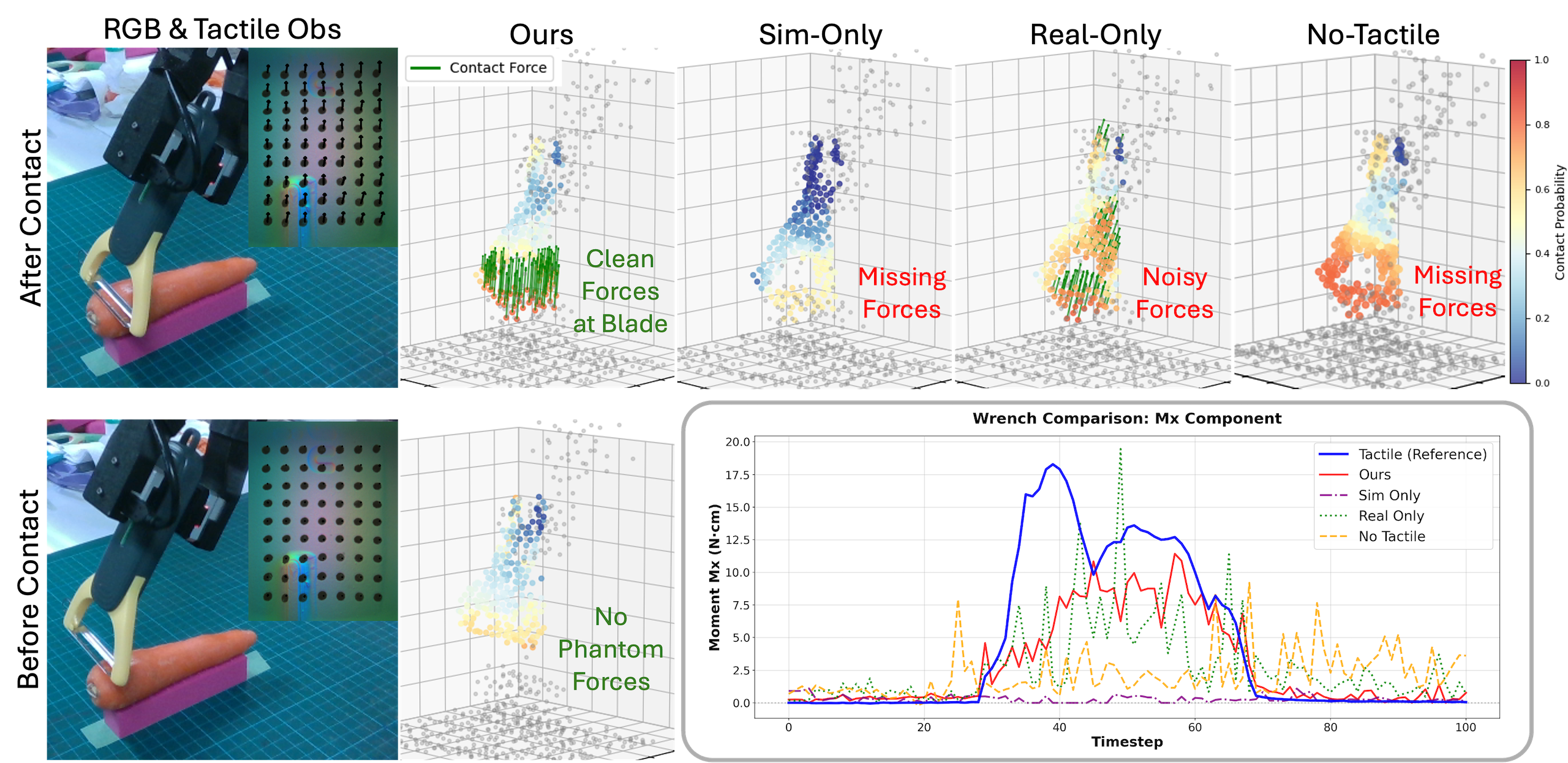}
    \caption{Qualitative comparison of contact-field estimation on the Peeler. \textbf{Top Row}: Ours produces clean contact forces localized on the blade-carrot interface, while Sim-Only and No-Tactile miss forces, Real-Only predicts noisy forces. \textbf{Bottom Right}: Correlation between torque induced by predicted contact forces and the reference wrench from tactile signals. Ours aligns best with the reference wrench, while Real-Only and No-Tactile remain noisy.}
    \label{fig:qualitative_vis}
    \vspace{-10pt}
\end{figure*}


\subsubsection{Evaluation 3: Qualitative Generalization}
Finally, we qualitatively evaluate generalization to the \textbf{Peeler} task, where complex interactions with irregular carrots make heuristic ground-truth labeling unreliable. For this task, the model is pre-trained on simulated peeler data but fine-tuned using \textbf{only} the real-world scraper dataset. As shown in Figure~\ref{fig:qualitative_vis}, Sim-Only misses real contact due to the tactile domain gap, Real-Only produces noisy forces without sufficient geometric priors, and No-Tactile fails to infer loaded blade contact from geometry alone. In contrast, Ours produces localized force estimates at the blade-carrot interface, suggesting that scraper-based real alignment can transfer to more complex curved tools when simulation provides the relevant contact prior.

\subsection{Policy Evaluation}
\label{subsec:exp_policy}

\begin{figure}[htbp]
    \centering
    \includegraphics[width=1\linewidth]{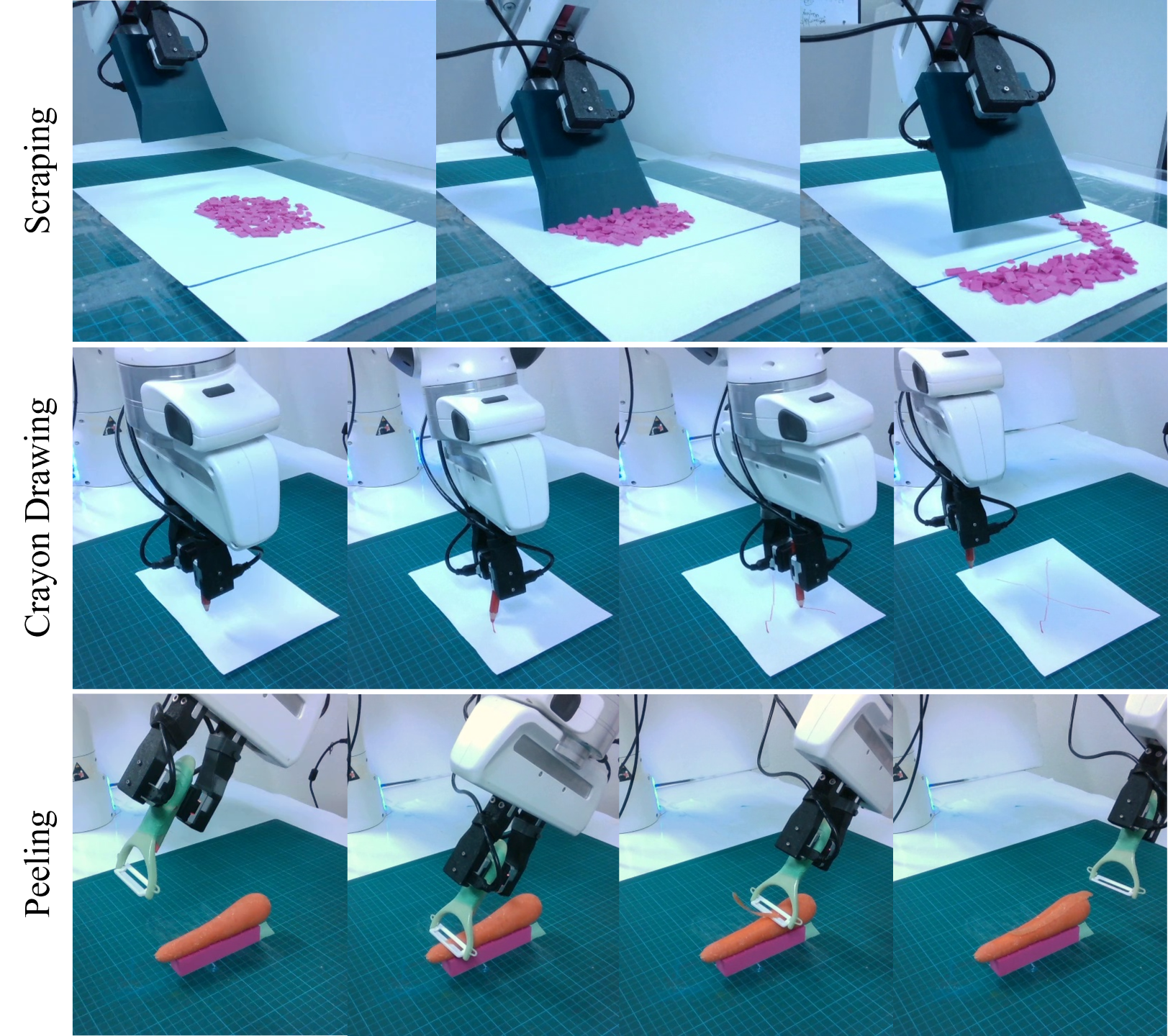}
    \caption{Rollouts of contact-rich tasks with unseen tools. \textbf{Top}: Scraping debris past a target line. \textbf{Middle}: Drawing a cross with consistent force. \textbf{Bottom}: Peeling a carrot. Additional experiment visualizations are available in Appendix \ref{app:additional_exp}.}
    \vspace{-15pt}
    \label{fig:combined_tasks}
\end{figure}

\noindent\textbf{Experimental Setup.} We evaluate policy performance on a real-world Franka Emika Panda robot. The robot is equipped with a parallel gripper modified to house two GelSight Mini tactile sensors. Visual observations are captured via three calibrated RealSense D435 cameras (front, left, right). We evaluate the Diffusion Policy conditioned on SCFields across three contact-rich tasks (Figure \ref{fig:combined_tasks}).

\vspace{3pt}

\subsubsection{Baselines and Ablations}
We compare SCFields against the \textbf{Vision-Only} baseline based on GenDP \cite{wang2024gendp}, and a \textbf{Raw Tactile (End-to-End)} baseline, which concatenates raw tactile data directly into the policy observation without explicit physics supervision.
Additionally, we evaluate the \textbf{Sim-Only Contact Field} and \textbf{Real-Only Contact Field} baselines defined in Section \ref{subsec:exp_contact_field}. Finally, a \textbf{No Contact Force} ablation isolates explicit contact force vectors by training the policy using only contact probability.

\vspace{3pt}

\subsubsection{Tasks and Metrics}

\begin{table*}[h]
    \centering
    \caption{Task 1: Scraper Performance. Ours outperforms baselines on unseen tools, demonstrating robust generalization.}
    \label{tab:scraper_results}
    \small
    \begin{tabular}{l|ccc|ccc}
        \toprule
        & \multicolumn{3}{c|}{\textbf{Seen Tools}} & \multicolumn{3}{c}{\textbf{Unseen Tools}} \\
        Method & SR (\%) & Eff (\%) & Eff Norm (\%) & SR (\%) & Eff (\%) & Eff Norm (\%) \\
        \midrule
        Vision-Only (GenDP) & 39.1 & 10.1 & 30.5 & 35.1 & 25.4 & 35.1 \\
        Raw Tactile & 35.1 & 26.9 & 35.7 & 50.0 & 23.3 & 27.3 \\
        \textbf{Ours (SCFields)} & \textbf{73.5} & \textbf{61.8} & \textbf{85.2} & \textbf{79.6} & \textbf{73.5} & \textbf{84.7} \\
        \midrule
        Ablation: Sim-Only CF & 34.5 & 20.6 & 33.1 & 55.6 & 37.0 & 45.2 \\
        Ablation: Real-Only CF & 45.8 & 32.3 & 44.5 & 50.0 & 47.5 & 54.2 \\
        Ablation: No Force & 31.3 & 22.2 & 29.9 & 26.0 & 24.3 & 27.6\\
        \bottomrule
    \end{tabular}
    \vspace{-10pt}
\end{table*}

\begin{table}[h]
    \centering
    \caption{Task 2: Crayon Drawing Consistency (Score 0-1).}
    \label{tab:crayon_results}
    \small
    \begin{tabular}{lcc}
        \toprule
        Method & \textbf{Seen Crayons} & \textbf{Unseen Crayons} \\
        \midrule
        Vision-Only (GenDP) & 0.81 & 0.60 \\
        Raw Tactile & 0.76 & 0.61 \\
        \textbf{Ours (SCFields)} & \textbf{0.86} & \textbf{0.78} \\
        \midrule
        Ablation: Sim-Only CF & 0.81 & 0.60 \\ 
        Ablation: Real-Only CF & 0.68 & 0.74 \\
        Ablation: No Force & 0.80 & 0.76 \\ 
        \bottomrule
    \end{tabular}
    \vspace{-15pt}
\end{table}

\begin{table*}[h]
    \centering
    \caption{Task 3: Peeler Results. Aligned model performance validates the pipeline.}
    \label{tab:peeler_results}
    \small
    \begin{tabular}{l|ccc|ccc}
        \toprule
        & \multicolumn{3}{c|}{\textbf{Seen Peelers}} & \multicolumn{3}{c}{\textbf{Unseen Peelers}} \\
        Method & Contact (\%) & Cut-in (\%) & Avg Peel Length (cm) & Contact (\%) & Cut-in (\%) & Avg Peel Length (cm) \\
        \midrule
        Vision-Only & 45.0 & 30.0 & 1.50 & 50.0 & 33.3 & 1.12 \\
        Raw Tactile & 45.0 & 20.0 & 1.05 & 40.0 & 30.0 & 0.85\\
        \textbf{Ours (SCField)} & \textbf{80.0} & \textbf{70.0} & \textbf{4.73} & \textbf{90.0} & \textbf{73.3} & \textbf{4.52} \\
        \midrule
        Ablation: Sim-Only CF & 60.0 & 30.0 & 2.00 & 50.0 & 46.7 & 3.05 \\
        Ablation: Real-Only CF & 60.0 & 15.0 & 0.93 & 57.5 & 50.0 & 1.80 \\
        Ablation: No Force & 65.0 & 30.0 & 1.95 & 40.0 & 13.3 & 1.08 \\
        \bottomrule
    \end{tabular}
    \vspace{-5pt}
\end{table*}

\noindent\textbf{Task 1: Scraper (Contact-Rich Cleaning).} The robot must maintain surface contact to clean debris. We trained the policy on 150 demonstration episodes collected across 3 table heights using 4 training tools. We evaluate on both seen and 4 unseen tools with 2 unseen table heights and 2 trials each (16 trials in total per split); dynamic stopping criteria yield 46--72 individual scrape attempts per split. Metrics include \textit{Success Rate (SR)} (\% of trials maintaining contact) and \textit{Cleaning Efficiency (Eff)} (\% of debris removed). An additional \textit{Normalized Efficiency (Eff Norm)} is included to offset the effect of different scraper blade lengths.

\noindent\textbf{Task 2: Crayon Drawing.} Picking up an asymmetric crayon (or pencil) to draw a cross. The policy was trained on 120 episodes across 3 heights using 3 training crayons and was evaluated on 2 unseen heights and 3 trials each (18 trials in total per split). Success requires precise force modulation to leave a visible trace without snapping the crayon. Metric: \textit{Drawing Consistency} (0-1 score reflecting completeness/visibility). The preliminary stage of picking up the crayon, which requires visual semantics generalization to unseen crayons/pencils, is reported in Appendix \ref{app:crayon_picking}.

\noindent\textbf{Task 3: Peeler.} Peeling a carrot with a handheld peeler. We trained the policy on 60 demonstration episodes using 2 training peelers and evaluated on 20 seen-peeler trials and 30 unseen-peeler trials. This task tests the cross-category generalization of our perception module, as the contact field model was aligned using only scraper data. Metrics: Percentage of Successful \textit{Contact} and \textit{Cut-in}, and \textit{Peel Quality} (avg. peel length).

\begin{figure}[h]
    \centering
    \includegraphics[width=1\linewidth]{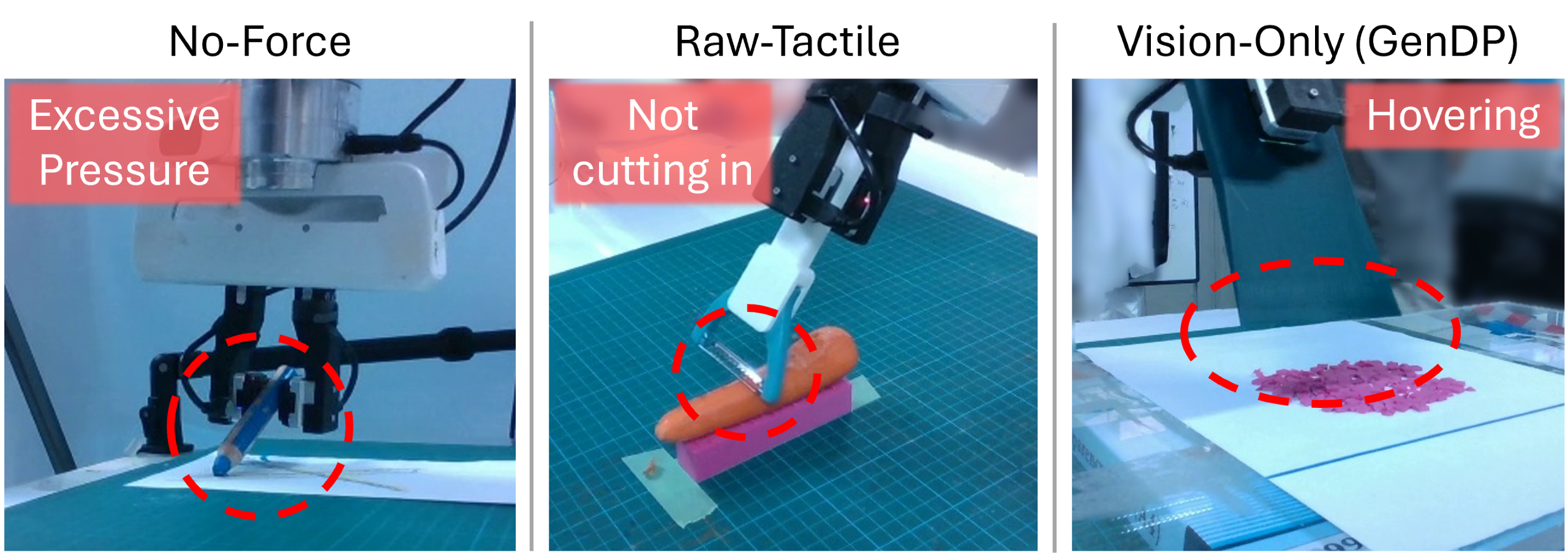}
    \caption{Example failure modes of baseline/ablation methods.}
    \label{fig:failures}
    \vspace{-15pt}
\end{figure}

\subsubsection{Analysis}
We present the quantitative results for all three tasks in Tables \ref{tab:scraper_results}, \ref{tab:crayon_results}, and \ref{tab:peeler_results}. Full experiment statistics including confidence intervals and $p$-values are reported in Appendix~\ref{app:statistics}.

\noindent\textbf{Comparison with Baselines.}
Across all tasks (Tables \ref{tab:scraper_results}, \ref{tab:crayon_results}, \ref{tab:peeler_results}), SCFields significantly outperforms baselines, particularly on unseen tools. The \textit{Raw Tactile} baseline struggles (e.g., 23.3\% Eff on Unseen Scraper), confirming that without explicit physical grounding, end-to-end policies fail to leverage high-dimensional tactile data effectively, often overfitting to visual inputs. Similarly, the \textit{Sim-Only CF} baseline generally matches Vision-Only performance, underscoring that without our real-world alignment stage, the domain gap renders tactile predictions unreliable. Conversely, while the \textit{Real-Only CF} ablation achieves better performance than other baselines on the seen scraper task, it fails to generalize to novel tools, confirming that simulation pre-training is requisite for learning contact representations that transfer across object categories.

\noindent\textbf{Role of Explicit Force \& Generalization.}
The \textbf{No Force} ablation isolates the value of continuous force prediction. Contact probability alone can indicate likely interaction regions, but it cannot distinguish insufficient loading from excessive pressure. As a result, the policy exhibits failure modes such as hovering above the surface or pressing too hard and slipping, showing that force vectors are necessary for regulating interaction dynamics.
Similar failure modes are also present in other baseline methods that do not explicitly model contact force, as illustrated in Figure \ref{fig:failures}. Notably, the Peeler task highlights the robustness of our pipeline: although the policy used imitation learning on peeling demonstrations, the underlying perception model was aligned using only scraper data. Despite this, our model achieves an average peel length of \textbf{4.52cm}, quadrupling the Vision-Only baseline (\textbf{1.12cm}). This confirms that SCFields successfully transfers the invariant concept of "functional contact" from simulation to novel real-world tools.

\section{Conclusion}
\label{sec:conclusion}

In this work, we presented \textbf{Semantic-Contact Fields (SCFields)}, a novel 3D representation that fuses visual semantics with dense, physically-grounded contact estimates to enable category-level generalization of contact-rich tool manipulation. We addressed the fundamental challenge of tactile sim-to-real transfer through a two-stage learning pipeline: pre-training on large-scale physics simulations to learn geometric-aware contact priors, followed by a data-efficient real-world alignment stage. This approach effectively bridges the reality gap without expensive instrumentation or high-fidelity simulation. Experiments on scraping, drawing, and peeling demonstrate that SCFields significantly outperforms baselines; by grounding sparse tactile readings into dense physical estimates, our system achieves zero-shot generalization to novel tool variants and robustness to dynamic environments where traditional methods fail.

A key limitation of the current framework is its reliance on imitation learning, which restricts the robot to tool usage patterns present in the demonstrations. While SCFields enable robust generalization across geometric variants within a category, the system cannot currently discover novel functional affordances or alternative ways of using tools—such as repurposing a knife to peel a carrot. Exploring Reinforcement Learning or World Models to enable the autonomous discovery of such creative tool manipulation strategies would be a valuable future direction.
Beyond these policy-level limitations, SCFields also inherits two system-level limitations. First, it assumes that the functional contact region is at least partially observable in the tool point cloud. Second, our current setup uses multiple RGB-D cameras and tactile sensors to obtain reliable tool geometry and contact estimates under gripper self-occlusion. Reducing sensing complexity and temporal tracking for fully occluded contact states remain important future work.

\bibliographystyle{plainnat}
\bibliography{references}

\newpage
\onecolumn
\newpage
\appendix
\setcounter{section}{0}

\tcbset{
  onecol/.style={
    colback=gray!10,
    colframe=gray!80,
    boxrule=0.5pt,
    arc=2mm,
    fontupper=\scriptsize\ttfamily,
    left=1mm,
    right=1mm,
    top=1mm,
    bottom=1mm,
    boxsep=1mm,
    breakable,
    width=\linewidth,
  }
}

\tcbset{
  twocol/.style={
    colback=gray!10,
    colframe=gray!80,
    boxrule=0.5pt,
    arc=2mm,
    fontupper=\scriptsize\ttfamily,
    left=1mm,
    right=1mm,
    top=1mm,
    bottom=1mm,
    boxsep=1mm,
    nobeforeafter,
    breakable,
    width=(\linewidth-2mm)/2,
  }
}

\subsection{Network Architecture Details}
\label{app:architecture}

We utilize a PointNet++ \cite{qi2017pointnet++} architecture to process the heterogeneous input point cloud. The network consists of a series of Set Abstraction (SA) layers for feature downsampling and Feature Propagation (FP) layers for upsampling.

\noindent\textbf{Input:} The input is a point cloud of size $(N, 3 + C_{in})$, where $N=894$ (256 object + 512 environment + 126 tactile) and $C_{in}=16$ (1 type channel + 15 tactile history channels).

\noindent\textbf{Encoder (Set Abstraction):}
\begin{itemize}
    \item \textbf{SA1:} Number of points: 512, Radius: 0.02m, Samples: 32, MLP: [32, 32, 64].
    \item \textbf{SA2:} Number of points: 128, Radius: 0.04m, Samples: 64, MLP: [64, 64, 128].
    \item \textbf{SA3 (Global):} Number of points: None (Global pooling), MLP: [128, 128, 256].
\end{itemize}

\noindent\textbf{Decoder (Feature Propagation):}
\begin{itemize}
    \item \textbf{FP1:} Interpolates features from SA3 to SA2. MLP: [256, 256].
    \item \textbf{FP2:} Interpolates features from FP1 to SA1. MLP: [256, 128].
    \item \textbf{FP3:} Interpolates features from FP2 to the original input points. MLP: [128, 128, 128].
\end{itemize}

\noindent\textbf{Prediction Heads:}
The decoded features (dim 128) are passed to two parallel heads:
\begin{enumerate}
    \item \textbf{Contact Probability Head:} MLP: [64, 32, 1] followed by a Sigmoid activation.
    \item \textbf{Force Regression Head:} MLP: [64, 32, 3] (No activation).
\end{enumerate}

\subsection{Simulation and Data Labeling Details}
\label{app:sim_data_labeling}

As described in Section \ref{subsubsec:sim_data_generation}, we employ a two-stage replay process to generate dense ground-truth labels from rigid-body simulation data. This section details the simulation configuration and the mathematical formulation used to map discrete rigid-body states to dense contact fields.

\begin{figure*}[h]
    \centering
    \includegraphics[width=0.9\linewidth]{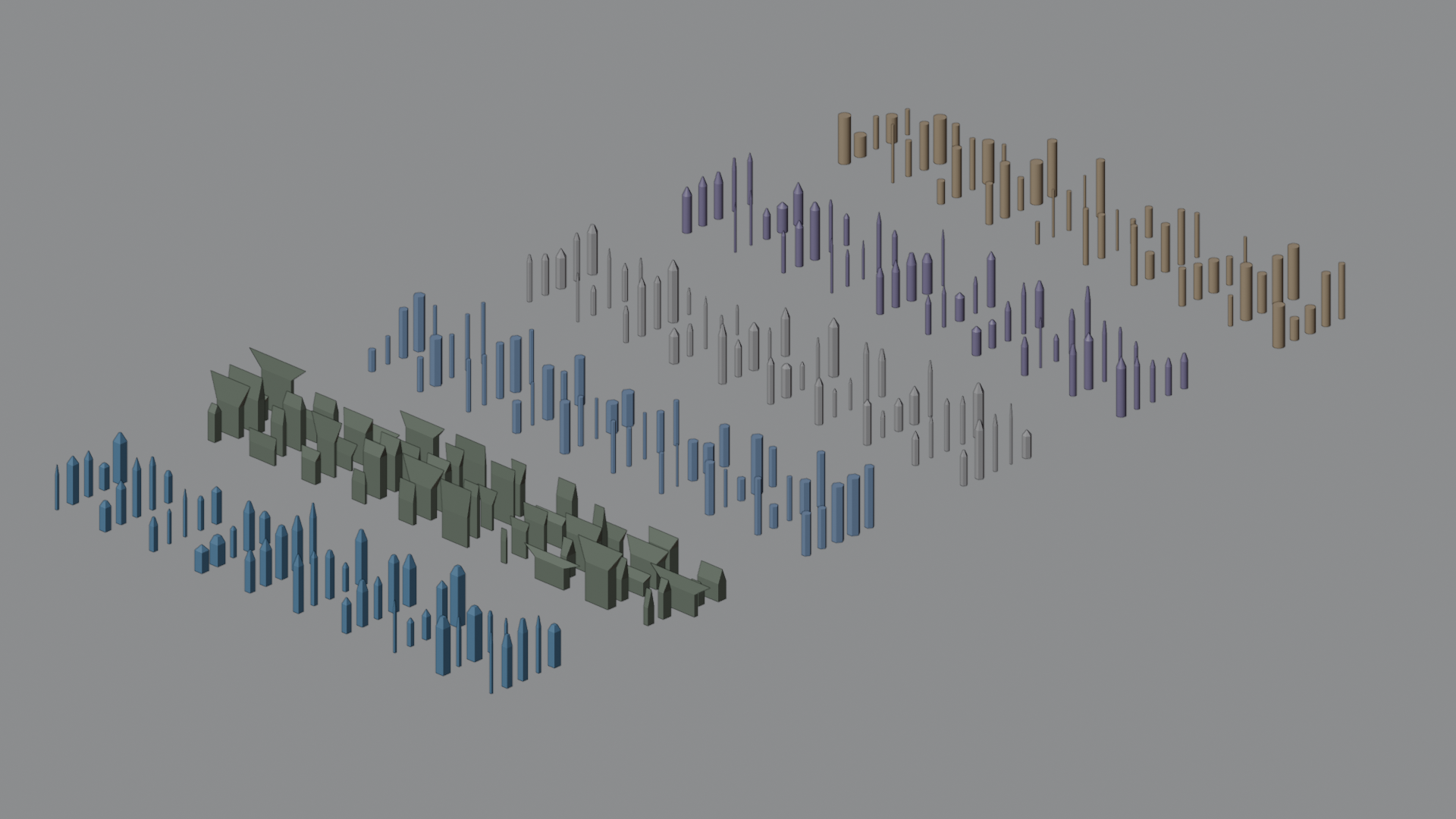}
    \label{fig:tool_meshes}
\end{figure*}

\begin{figure*}[h]
    \centering
    \includegraphics[width=0.9\linewidth]{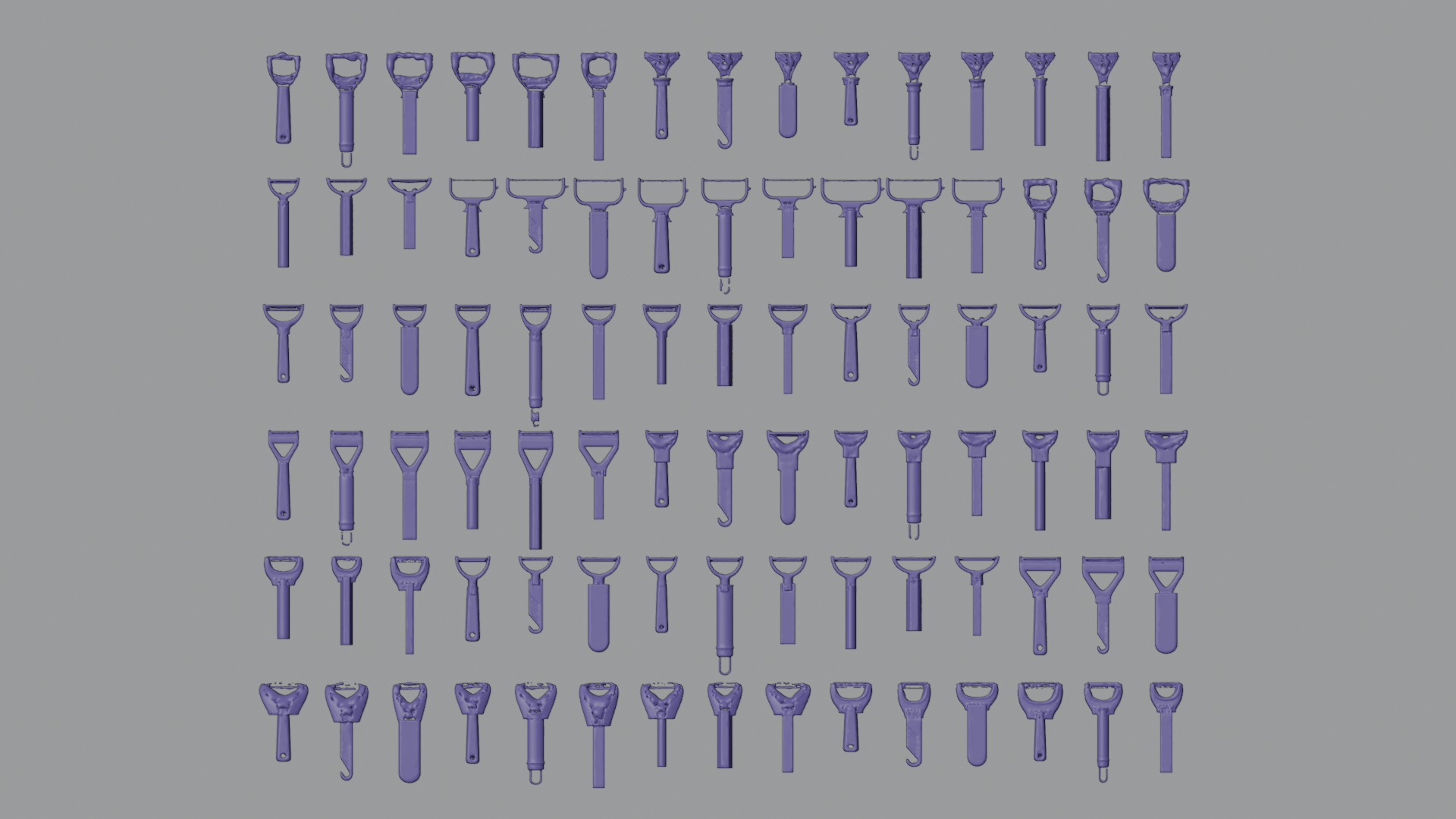}
    \caption{Peeler meshes used in simulation}
    \label{fig:peeler_meshes}
    \vspace{-5pt}
\end{figure*}

\subsubsection{Simulation Environments and Tactile Modeling}
Our simulation pipeline utilizes the TacSL framework~\cite{akinola2025tacsl} to model the physics of the GelSight sensor within IsaacGym \cite{makoviychuk2021isaac}. We define a uniform $7 \times 9$ marker grid that matches the physical distribution of the GelSight Mini sensors used in our real-world experiments. 

We employ TacSL's penalty-based tactile model to derive shear force distributions and surface depth maps at each marker location. This point-based representation serves as a transferable input for our architecture, maintaining consistency across both simulated and real tactile data streams.

\noindent \textbf{Simulation Configuration}
We provide the specific physical and control parameters used in our IsaacGym and TacSL setup in Listing~\ref{lst:sim_params}. The compliance parameters were randomized during training to improve the robustness of the learned policies.

\begin{figure}[h]
    \centering
    \begin{minipage}{0.48\linewidth}
        \begin{lstlisting}[language=yaml, caption={Key Simulation and TacSL parameters.}, label={lst:sim_params}, frame=single, basicstyle=\ttfamily\small]
TacSL:
  compliance_stiffness_range: [1400, 1500]
  compliant_damping_range: [1.5, 2.5]
  elastomer_friction: 5.0

IsaacGym:
  substeps: 4
  physx:
    num_pos_iterations: 32
    num_vel_iterations: 2
    contact_offset: 0.002
    max_depenetration_vel: 1.0
    friction_corr_dist: 0.001

Robot_Control:
gripper_prop_gains: [800, 800]
gripper_deriv_gains: [4, 4]
task_space_impedance: # cartesian
  prop_gains: [800, 800, 600, 100, 100, 100]
  deriv_gains: [100, 100, 75, 3, 3, 3]
  # prop_gains randomization range
  kp_min: [300, 300, 300, 20, 20, 20]
  kp_max: [800, 800, 800, 60, 60, 60]
        \end{lstlisting}
    \end{minipage}
    \hfill
    \begin{minipage}{0.48\linewidth}
\begin{lstlisting}[language=yaml, caption={Tactile filtering and smoothing parameters.}, label={lst:smoothing_params}, frame=single, basicstyle=\ttfamily\small]
Tactile_Filtering:
  spatial:
    enabled: true
    method: "gaussian"
    sigma: 0.25
  temporal:
    enabled: true
    window_length: 7
    polyorder: 1

Contact_Smoothing:
  precontact_smoothing: true
  postcontact_smoothing: true
  method: "linear"
  depth_threshold: -0.002
\end{lstlisting}
    \end{minipage}
\vspace{-10pt}
\end{figure}

\subsubsection{Tactile Data Post-Processing}To improve the quality of the tactile signal, we apply a multi-stage post-processing pipeline to the raw simulated tactile data. This includes spatial filtering to emulate the elastic diffusion of the elastomer, temporal filtering to reduce simulation jitter, and contact-phase smoothing to ensure a clean baseline.

\noindent \textbf{Spatial and Temporal Filtering} We apply a spatial Gaussian filter to the $7 \times 9$ marker grid to simulate the physical coupling between adjacent taxels in a real elastomer. Additionally, a temporal Savitzky-Golay filter is applied across a sliding window of timesteps to suppress high-frequency noise inherent in the physics solver's penalty-based contact model.

\noindent \textbf{Contact-Phase Smoothing} A significant challenge in simulated tactile data is the presence of non-zero force residuals when the sensor is not in contact. To address this, we implement a phase-aware smoothing strategy. Based on the ground-truth contact depth, we identify the \textit{pre-contact} (approach) and \textit{post-contact} (lifting) phases. As detailed in Listing~\ref{lst:smoothing_params}, we apply a linear interpolation from the phase median to the boundary contact value, effectively neutralizing sensor drift and simulation artifacts during non-contact states.

We observe that tactile simulation fidelity is highly sensitive to physical parameters. Despite rigorous tuning and data filtering, the quality of simulated tactile signals remains limited, as detailed in Section \ref{subsec:exp_contact_field}. This persistent discrepancy underscores the critical necessity of our real-world alignment stage to effectively bridge the sim-to-real gap.

\subsubsection{Soft Contact Probability Labeling}
Rigid-body simulators typically treat contact as a binary and unstable state. To generate smooth, learnable contact probability labels, we utilize the Signed Distance Function (SDF) computed in Open3D. We map the penetration depth $d_i$ (where $d_i < 0$ indicates penetration) of each point $p_i$ on the tool surface to a continuous contact probability $c_i \in [0, 1]$ using a one-sided exponential decay function:
\begin{equation}
    c_i = P(contact | d_i) = \exp \left( - \left( \frac{\max(-d_i, 0)}{\lambda} \right)^k \right)
\end{equation}
where $k=1.7$ controls the sharpness of the boundary, providing a smooth, Gaussian-like falloff. The length-scale parameter $\lambda$ is computed automatically such that the probability decays to $0.5$ at a penetration depth of $5\text{mm}$. This formulation ensures that points deep inside the object (indicating strong contact) have probabilities near 1.0, while points merely grazing the surface are assigned intermediate values.

\subsubsection{Dense Force Labeling by Extrapolation}
PyBullet provides discrete contact manifolds consisting of a sparse set of contact positions $\{\mathbf{x}_j\}$, normal vectors $\{\mathbf{n}_j\}$, and force magnitudes $\{F_j\}$. To transform these sparse interactions into a dense force field $\mathbf{f}^{ext}_i$ defined over the tool's point cloud, we employ a distance-weighted kernel interpolation modulated by local geometry.

For every point $p_i$ on the tool mesh, the extrapolated force vector is computed as:
\begin{equation}
    \mathbf{f}^{ext}_i = S(d_i) \cdot \frac{\sum_j w_{ij} (F_j \mathbf{n}_j)}{\sum_j w_{ij}}
\end{equation}

\noindent\textbf{Distance Weighting ($w_{ij}$):} We determine the influence of a discrete contact $j$ on mesh point $i$ using an inverse-square kernel based on their Euclidean distance:
\begin{equation}
    w_{ij} = \frac{1}{1 + (\lambda_{dist} \|\mathbf{x}_j - p_i\|)^2}
\end{equation}
where $\lambda_{dist}=50.0$ controls the locality, ensuring forces are concentrated around the active contact region.

\noindent\textbf{Depth Modulation:} To obtain smoother force distribution and prevent non-contacting points accumulate large amount of forces due to kernel interpolation, we scale the extrapolated force by the point's local penetration depth:

\begin{equation}
    S(d_i) = \sqrt{\text{ReLU}\left(1 - \frac{d_i}{d_{thresh}}\right)}
\end{equation}
where $d_{thresh}=-5\text{mm}$. This term ensures that the force magnitude tapers smoothly to zero as a point moves away from the penetration surface. Finally, we apply spatial outlier clipping (98th percentile) to remove numerical spikes inherent to rigid-body collision solving.

\subsubsection{Real-World Sensor Calibration}
To bridge the gap between simulated and real tactile readings, we perform a force calibration procedure on the real GelSight sensors. The calibration involves making the gripper grasp a reference block and then applying a known external force by placing a calibrated weight on the block. During this interaction, we monitor the change in the total wrench, which is computed from the tactile marker displacement and the depth map.

We calculate a scaling factor to align the magnitude of the computed tactile wrench with the known applied external wrench. It is important to note that this process does not aim to establish a precise, non-linear mapping from tactile signals to contact forces. Instead, it serves as a linear alignment step to ensure that the scale of the tactile signals in the real world matches the dynamic range observed in simulation, facilitating robust sim-to-real transfer.

\subsection{Training Hyperparameters and Loss Functions}
\label{app:training_params}

\subsubsection{Contact Field Loss Functions}
We optimize the network using a composite loss function:
\begin{equation}
    \mathcal{L}_{total} = \lambda_{prob} \mathcal{L}_{prob} + \lambda_{force} ( \lambda_{mag} \mathcal{L}_{mag} + \lambda_{dir} \mathcal{L}_{dir} )
\end{equation}
where $\lambda_{prob}=1.0$ and $\lambda_{force}=2.0$. Within the force term, the components are weighted by $\lambda_{mag}=1.5$ and $\lambda_{dir}=1.0$.

\noindent\textbf{Contact Probability Loss ($\mathcal{L}_{prob}$):} We use the Focal Loss to handle the extreme class imbalance (contact vs. free space):
\begin{equation}
    \mathcal{L}_{prob} = - \alpha_{t} (1-p_t)^\gamma \log(p_t)
\end{equation}
where $p_t$ is the model's estimated probability for the true class, with focusing parameter $\gamma=0.75$ and balancing factor $\alpha=0.9$ for the positive class.

\noindent\textbf{Force Magnitude Loss ($\mathcal{L}_{mag}$):} We use a Mean Squared Error (MSE) loss on the force norms, with adaptive weighting to prioritize high-force regions:
\begin{equation}
    \mathcal{L}_{mag} = w_i ( \|\mathbf{f}^{pred}\| - \|\mathbf{f}^{gt}\| )^2
\end{equation}
where $w_i$ is a log-magnitude adaptive weight $w_i \propto \log(1 + \|\mathbf{f}^{gt}\|)$ clipped to $[1.0, 3.0]$.

\noindent\textbf{Force Direction Loss ($\mathcal{L}_{dir}$):} We use Cosine Similarity loss, applied only to points where the ground truth force magnitude exceeds a threshold $\tau=0.005N$:
\begin{equation}
    \mathcal{L}_{dir} = 1 - \frac{\mathbf{f}^{pred} \cdot \mathbf{f}^{gt}}{\max(\|\mathbf{f}^{pred}\| \|\mathbf{f}^{gt}\|, \epsilon)}
\end{equation}

\subsubsection{Contact Field Training Schedule}
Table \ref{tab:hyperparameters} summarizes the training parameters.

\begin{table}[h]
    \centering
    \small
    \caption{Contact Field Model Training Hyperparameters}
    \label{tab:hyperparameters}
    \begin{tabular}{lcc}
        \toprule
        \textbf{Parameter} & \textbf{Stage 1 (Sim)} & \textbf{Stage 2 (Real)} \\
        \midrule
        Optimizer & AdamW & AdamW \\
        Learning Rate & $1e^{-4}$ & $5e^{-6}$ \\
        LR Scheduler & ReduceLROnPlateau & ReduceLROnPlateau \\
        Batch Size & 320 & 128 \\
        Epochs & 400 & 60 \\
        Point Translation & $\pm 0.1$m & $\pm 0.05$m \\
        Point Rotation & $\pm 30^{\circ}$ (Z-axis) & $\pm 15^{\circ}$ (Z-axis) \\
        Jitter Noise ($\sigma$) & 0.01 & 0.01 \\
        Tactile Noise ($\sigma$) & 0.001 & 0.001 \\
        \bottomrule
    \end{tabular}
\end{table}

\subsubsection{Diffusion Policy Hyperparameters}
We utilize a Diffusion Policy modeled as a conditional U-Net to predict robot actions. The policy takes a history of $T_{obs}=3$ observations and predicts a sequence of action steps with a prediction horizon of $T=16$, executing $T_{action}=8$ steps before replanning. The specific hyperparameters are detailed in Table \ref{tab:diffusion_params}.

\begin{table}[h]
    \centering
    \small
    \caption{Diffusion Policy Hyperparameters}
    \label{tab:diffusion_params}
    \begin{tabular}{ll}
        \toprule
        \textbf{Parameter} & \textbf{Value} \\
        \midrule
        \multicolumn{2}{l}{\textit{Policy Configuration}} \\
        Observation Horizon ($T_{obs}$) & 3 \\
        Action Execution ($T_{action}$) & 8 \\
        Prediction Horizon ($T$) & 16 \\
        Action Type & Delta End-Effector Pose \\
        \midrule
        \multicolumn{2}{l}{\textit{Optimization}} \\
        Optimizer & AdamW \\
        Learning Rate & $1e^{-4}$ \\
        Weight Decay & $1e^{-6}$ \\
        LR Scheduler & Cosine w/ 500 warmup steps \\
        Batch Size & 128 \\
        Epochs & 1000 \\
        EMA & Enabled (Power 0.75) \\
        \bottomrule
    \end{tabular}
\end{table}

\subsection{Crayon Picking Experiment}
\label{app:crayon_picking}

\begin{figure*}[htbp]
    \centering
    \includegraphics[width=0.9\linewidth]{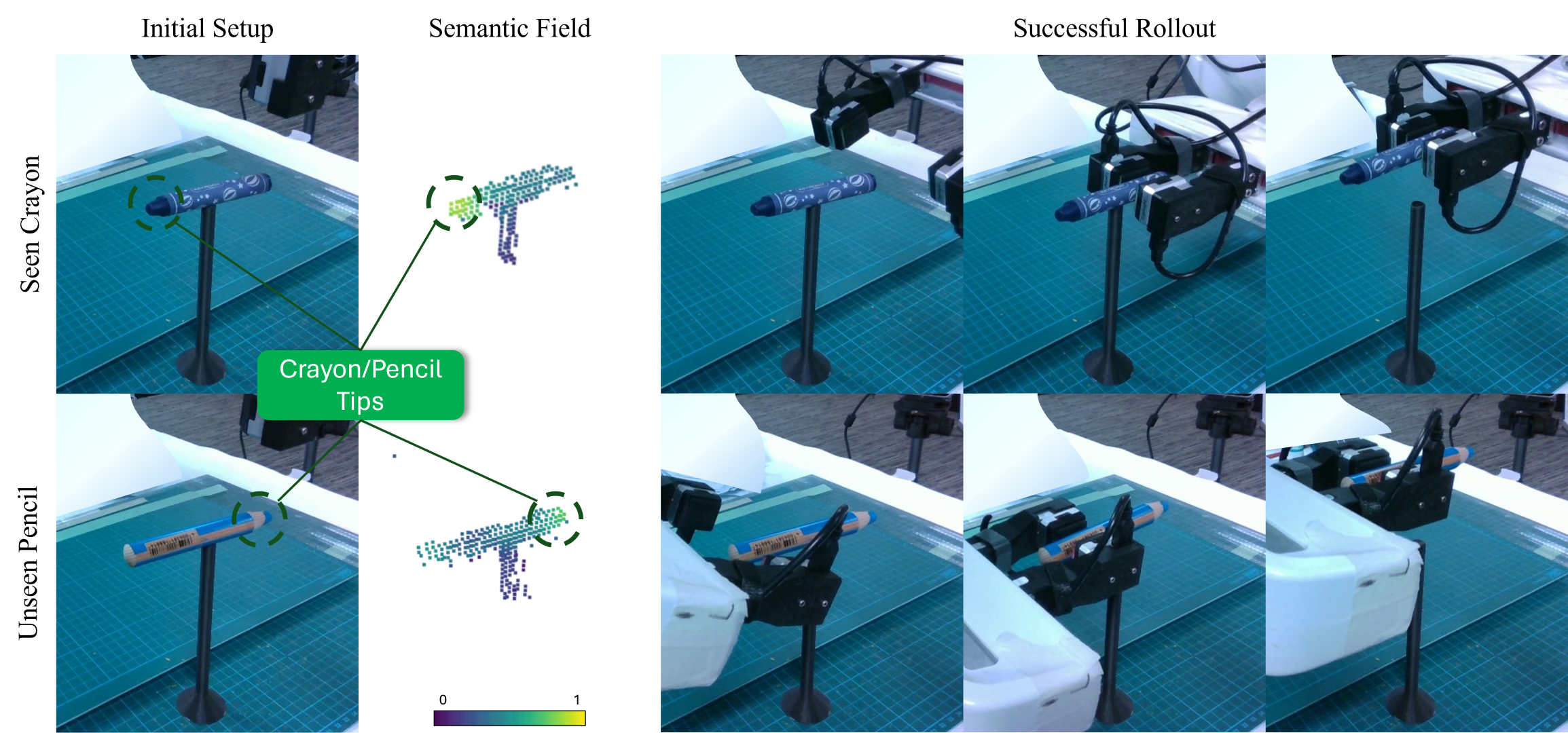}
    \caption{Illustration of the crayon picking setup, Semantic Field visualization, and successful rollouts on both seen and unseen instances. The Semantic Field is able to distinguish between the tip and handle in both seen and unseen crayons/pencils, guiding the robot to pick up from the correct direction.}
    \label{fig:crayon_picking_example}
\end{figure*}

Although not the primary focus of this work, we conducted a crayon picking experiment to evaluate the utility of the semantic field as a prerequisite capability for the drawing task. The experiment entails identifying and grasping a crayon or pencil placed on a holder, as illustrated in Figure~\ref{fig:crayon_picking_example}. 

Given the asymmetric geometry of the tools, the robot must explicitly grasp the handle rather than the writing tip to enable subsequent use. We evaluate our method against a baseline that excludes the semantic field, relying solely on the contact field, point cloud coordinates ($XYZ$), and $RGB$ information. Both models were trained for 60 epochs using the dataset collected from the three training crayons shown in Figure~\ref{fig:setup}. 

We assess performance using two metrics: \textit{Directional Accuracy} (the percentage of trials where the robot approaches the correct handle side) and \textit{Grasp Success Rate} (the percentage of successful lifts). As detailed in Table~\ref{tab:crayon_pick_results}, the inclusion of the semantic field significantly improves performance.

\begin{table}[htbp]
    \centering
    \caption{Crayon Picking Experimental Results. We report the Directional Accuracy (Dir. Acc.) and Grasp Success Rate (Success) on seen and unseen objects.}
    \label{tab:crayon_pick_results}
    \begin{tabular}{lcccc}
        \toprule
        & \multicolumn{2}{c}{Seen Objects} & \multicolumn{2}{c}{Unseen Objects} \\
        \cmidrule(lr){2-3} \cmidrule(lr){4-5}
        Method & Dir. Acc. & Success & Dir. Acc. & Success \\
        \midrule
        Baseline (w/o Semantic) & 63.3\% & 33.3\% & 40.0\% & 23.3\% \\
        \textbf{Ours (w/ Semantic)} & \textbf{93.3\%} & \textbf{76.7\%} & \textbf{93.3\%} & \textbf{73.3\%} \\
        \bottomrule
    \end{tabular}
\end{table}

The second column in Figure~\ref{fig:crayon_picking_example} qualitatively demonstrates that our model infers generalized semantic fields across both seen and unseen crayons/pencils, successfully highlighting the tip area to avoid. The quantitative results further confirm that the semantic field enables the policy to consistently identify the handle location and approach from the correct direction. Conversely, the baseline policy often fails to distinguish the handle from the tip, resulting in performance close to random guessing, particularly on unseen objects.

\subsection{Additional Experiment Details and Qualitative Analysis}
\label{app:additional_exp}

\subsubsection{Details on Evaluation Metrics for Scraping Task}
In the scraping task described in the main text, we employ two primary metrics to evaluate performance: Scraping Efficiency (\textit{Eff}) and Normalized Scraping Efficiency (\textit{Eff Norm}).

\begin{itemize}
    \item \textbf{Scraping Efficiency (\textit{Eff}):} This metric measures the percentage of debris successfully removed. We weigh the debris pushed behind the target line (the blue line) using a precision scale. The efficiency is defined as the ratio of cleaned weight to total weight:
    \begin{equation}
        \text{Eff} = \frac{W_{\text{cleaned}}}{W_{\text{total}}}
    \end{equation}

    \item \textbf{Normalized Scraping Efficiency (\textit{Eff Norm}):} Tools with longer blades naturally cover a larger area and tend to achieve higher raw scraping efficiency. Since the tools in our test set possess a longer average blade length than those in the training set, direct comparison using raw efficiency is biased. To account for this geometric advantage, we normalize the efficiency by the tool's blade length. We define a reference blade length ratio $L_{\text{ref}} = L_{\text{blade}} / L_{\text{max}}$, where $L_{\text{max}}$ is the length of the longest blade across all tools. The normalized efficiency is computed as:
    \begin{equation}
        \text{Eff Norm} = \min\left(1, \frac{\text{Eff}}{L_{\text{ref}}}\right)
    \end{equation}
    This metric rewards policies that maximize the utility of the available tool geometry.
\end{itemize}

\subsubsection{Qualitative Evaluation of Contact Field Prediction}
To evaluate the robustness of our contact field estimation, we provide qualitative comparisons between the model's predictions and the ground truth (or pseudo-ground truth) data across both simulated and real-world domains.

\begin{figure*}[htbp]
    \centering
    \includegraphics[width=0.9\linewidth]{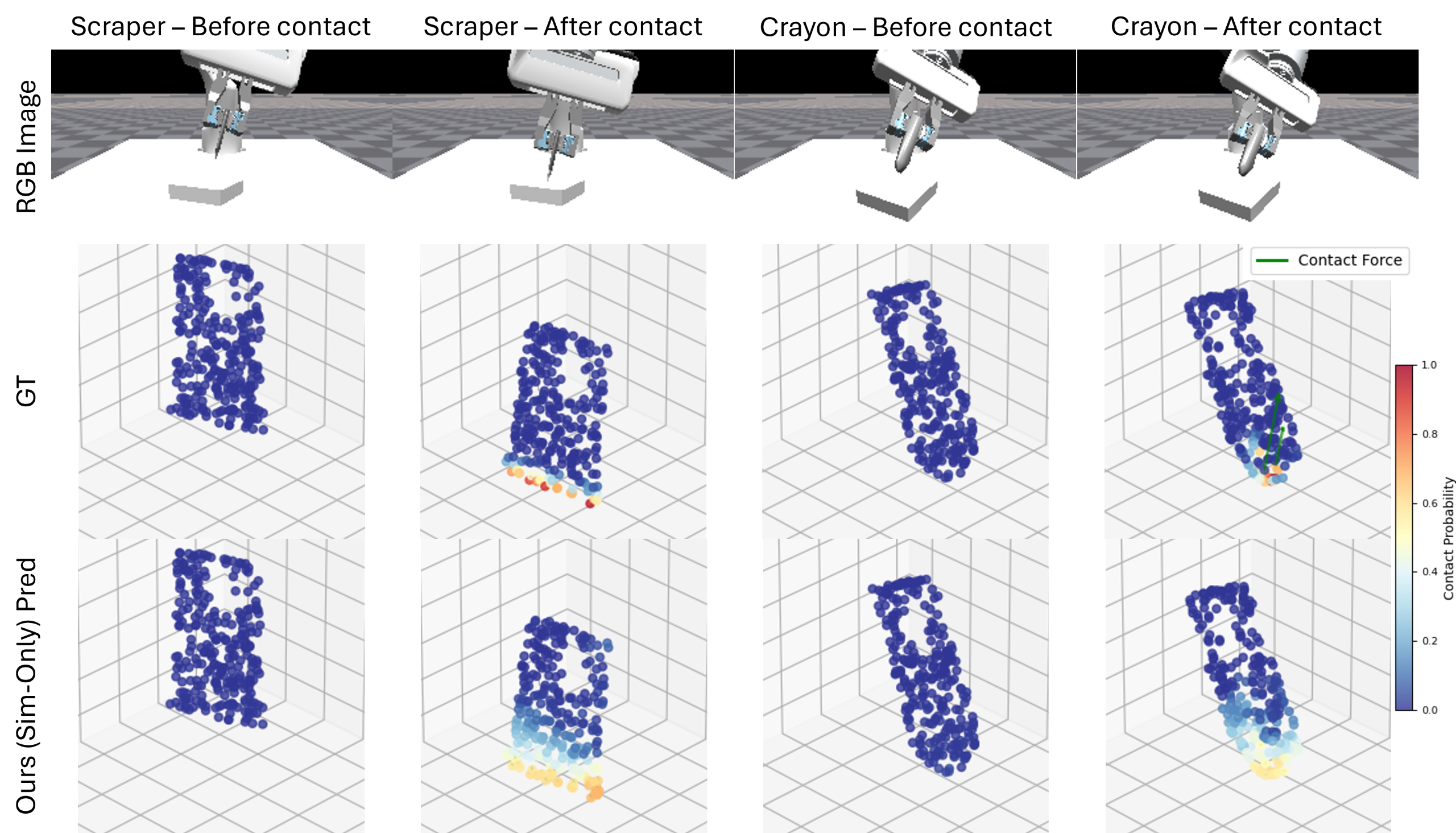}
    \caption{Qualitative results in simulation. The predicted contact probabilities (bottom row) closely match the ground truth fields (middle row) generated by the simulation pipeline.}
    \label{fig:sim_contact_vis}
    \vspace{-5pt}
\end{figure*}

\paragraph{Simulation Results}
Figure~\ref{fig:sim_contact_vis} illustrates the contact field prediction in the simulation environment. The model accurately reconstructs the contact geometry compared to the ground truth provided by the TacSL physics engine.

\begin{figure*}[htbp]
    \centering
    \includegraphics[width=0.9\linewidth]{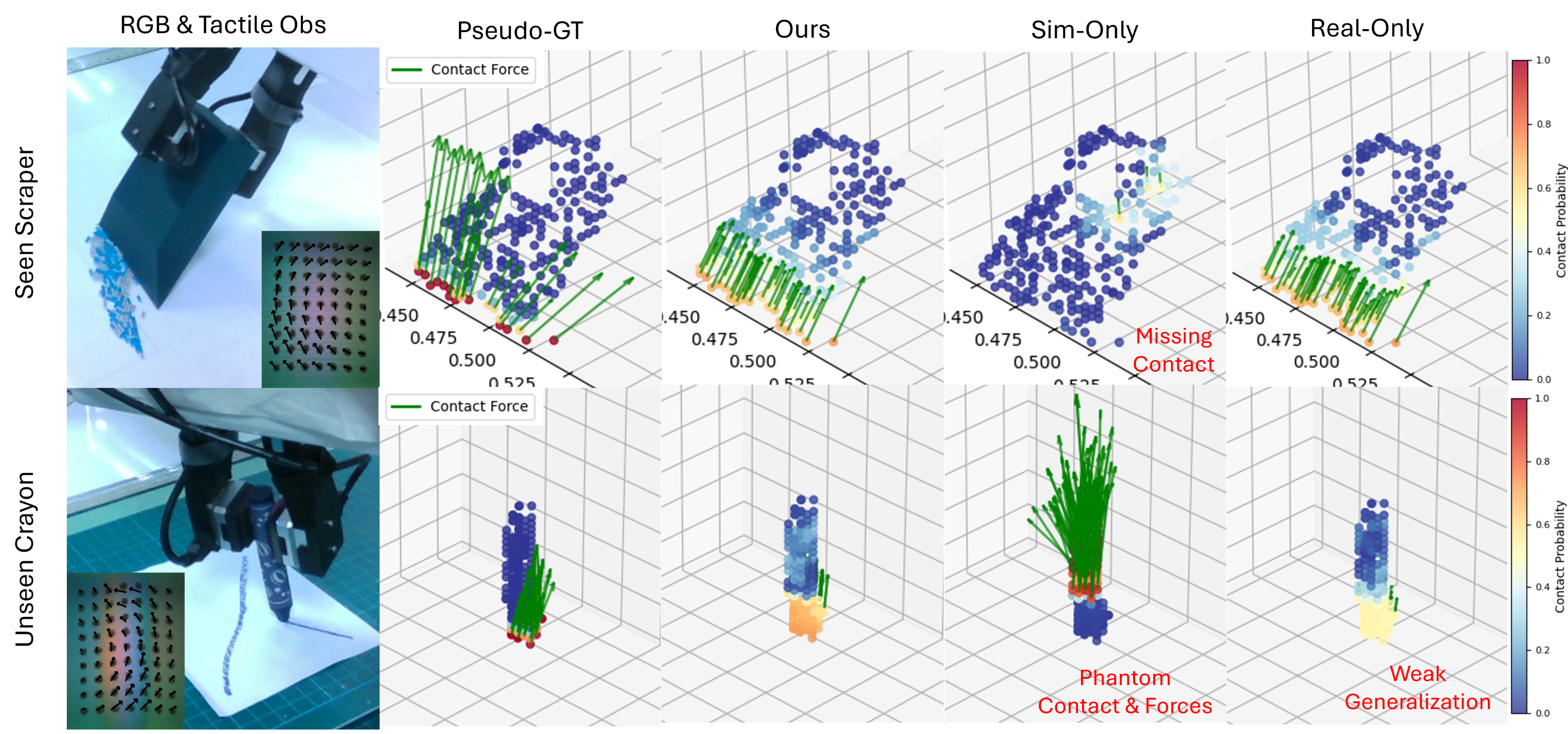}
    \caption{Qualitative results in the real world. We compare the predicted contact fields by \textbf{Ours}, \textbf{Sim-Only} baseline, and \textbf{Real-Only} baseline against pseudo-ground truth for the scraping tool (top) and the crayon (bottom). The Sim-Only baseline produces missing or phantom contact and forces. The Real-Only model performs well on scraper seen in training data, but is worse than ours in generalizing to the unseen crayon.}
    \label{fig:real_contact_vis}
    \vspace{-15pt}
\end{figure*}

\paragraph{Real-World Results}
In the real-world experiments, absolute ground truth for the contact field is unavailable. Instead, we generate a "pseudo-ground truth" derived from high-resolution depth maps captured by the GelSight sensor. Figure~\ref{fig:real_contact_vis} displays the predictions for both the scraping tool and the crayon grasping task. Despite the domain shift, the model successfully infers contact patches that align with the physical interaction areas.

\newpage

\FloatBarrier
\subsection{Statistical Analysis}
\label{app:statistics}
\suppressfloats[t]

We report additional variability and significance statistics for the contact-field estimator and real-robot policy evaluations. Table~\ref{tab:app_contact_field_stats} summarizes contact-field performance with confidence intervals and selected pairwise tests against Ours. Table~\ref{tab:app_scraper_stats} reports scraper success and blade-length-normalized cleaning efficiency. Table~\ref{tab:app_crayon_stats} reports crayon drawing consistency. Table~\ref{tab:app_peeler_stats} reports peeler contact, cut-in, and peel-length statistics. Unless otherwise stated, all intervals are 95\% confidence intervals. For binary policy metrics, we use Wilson score intervals and Fisher's exact test. For continuous policy metrics, we use bootstrap confidence intervals and non-parametric tests. For policy comparisons against Ours, we report raw $p$-values.

For the scraper task, Table~\ref{tab:app_scraper_stats} shows that SCFields improves contact-maintenance success over all baselines on both seen and unseen tools, and also improves blade-length-normalized cleaning efficiency.

For crayon drawing, Table~\ref{tab:app_crayon_stats} shows that SCFields achieves the highest average drawing consistency on both seen and unseen crayons. However, significance tests are not conclusive, so we interpret this task as supporting evidence rather than the strongest statistical result.

For peeling, Table~\ref{tab:app_peeler_stats} reports contact, cut-in, and peel-length statistics. These results provide the clearest evidence for the benefit of simulation-learned contact priors, especially on unseen peelers where dense real pseudo-labeling is difficult.

\begin{table*}[h]
\centering
\caption{\textbf{Contact-field evaluation with confidence intervals and selected pairwise tests.} F1 scores are computed per-frame to enable bootstrap significance testing, distinct from the aggregate scores in Table~\ref{tab:sim_metrics} and Table~\ref{tab:real_alignment}. No Contact Prob. does not output contact probability, so F1 is not applicable. $p$-values compare each method against Ours.}
\label{tab:app_contact_field_stats}
\resizebox{\textwidth}{!}{
\begin{tabular}{llcccc}
\toprule
Setting & Model 
& F1 Score $\uparrow$ 
& F1 $p$-value 
& Force MSE $\downarrow$ 
& Force MSE $p$-value \\
\midrule
\multirow{5}{*}{Sim}
& No-Tactile & 0.322 [0.317, 0.327] & $<$0.001 & 0.0146 [0.0141, 0.0151] & 1.000 \\
& 2D Tactile Encoder & 0.303 [0.298, 0.308] & $<$0.001 & 0.0147 [0.0142, 0.0152] & 0.011 \\
& BCE Loss & 0.022 [0.021, 0.024] & $<$0.001 & 0.0146 [0.0142, 0.0151] & 1.000 \\
& No Contact Prob. & N/A & N/A & 0.0158 [0.0153, 0.0162] & $<$0.001 \\
& \textbf{Ours} & \textbf{0.462 [0.457, 0.467]} & -- & 0.0147 [0.0142, 0.0151] & -- \\
\midrule
\multirow{5}{*}{Real Scraper}
& Sim-Only & 0.002 [0.001, 0.004] & $<$0.001 & 0.0435 [0.0408, 0.0463] & $<$0.001 \\
& Real-Only & 0.445 [0.435, 0.455] & $<$0.001 & \textbf{0.0221 [0.0204, 0.0238]} & 1.000 \\
& No-Tactile & 0.397 [0.385, 0.408] & $<$0.001 & 0.0432 [0.0406, 0.0458] & $<$0.001 \\
& No Contact Prob. & N/A & N/A & 0.0377 [0.0353, 0.0402] & $<$0.001 \\
& \textbf{Ours} & \textbf{0.518 [0.506, 0.529]} & -- & 0.0254 [0.0236, 0.0273] & -- \\
\midrule
\multirow{5}{*}{Real Crayon}
& Sim-Only & 0.005 [0.002, 0.008] & $<$0.001 & 0.0284 [0.0258, 0.0311] & $<$0.001 \\
& Real-Only & 0.485 [0.456, 0.513] & $<$0.001 & 0.0106 [0.0089, 0.0124] & 0.003 \\
& No-Tactile & 0.423 [0.395, 0.451] & $<$0.001 & 0.0115 [0.0089, 0.0143] & 0.004 \\
& No Contact Prob. & N/A & N/A & 0.0089 [0.0070, 0.0111] & 0.308 \\
& \textbf{Ours} & \textbf{0.524 [0.496, 0.552]} & -- & \textbf{0.0085 [0.0069, 0.0102]} & -- \\
\bottomrule
\end{tabular}
}
\end{table*}

\begin{table*}[h]
\centering
\caption{\textbf{Scraper policy evaluation statistics.} Success is computed over individual scrape attempts and reported as percentage with Wilson score intervals. Normalized cleaning efficiency is also reported as percentage. $p$ denotes the raw test $p$-value comparing each method against Ours.}
\label{tab:app_scraper_stats}
\resizebox{\textwidth}{!}{
\begin{tabular}{lcccccccc}
\toprule
\multirow{2}{*}{Method}
& \multicolumn{4}{c}{Seen Tools}
& \multicolumn{4}{c}{Unseen Tools} \\
\cmidrule(lr){2-5} \cmidrule(lr){6-9}
& Success [95\% CI] & $p$
& Eff. Norm. [95\% CI] & $p$
& Success [95\% CI] & $p$
& Eff. Norm. [95\% CI] & $p$ \\
\midrule
GenDP 
& 39.1 [26.4, 53.5] & $<$0.001
& 30.5 [13.3, 48.9] & $<$0.001
& 35.1 [24.0, 48.1] & $<$0.001
& 35.1 [21.2, 49.2] & $<$0.001 \\
Raw Tactile 
& 35.1 [24.0, 48.1] & $<$0.001
& 35.7 [16.0, 56.5] & 0.001
& 50.0 [36.4, 63.6] & 0.002
& 27.3 [10.2, 46.7] & $<$0.001 \\
Sim-Only CF 
& 34.6 [23.4, 47.8] & $<$0.001
& 33.1 [14.5, 53.6] & $<$0.001
& 55.6 [44.1, 66.5] & 0.004
& 45.2 [26.9, 64.3] & 0.001 \\
Real-Only CF 
& 45.8 [34.8, 57.3] & 0.001
& 44.5 [23.9, 65.6] & 0.003
& 50.0 [38.1, 61.9] & 0.001
& 54.2 [34.2, 73.3] & 0.009 \\
No Force 
& 31.3 [21.2, 43.4] & $<$0.001
& 29.9 [12.8, 49.0] & $<$0.001
& 26.0 [15.9, 39.6] & $<$0.001
& 27.6 [11.4, 45.7] & $<$0.001 \\
\textbf{Ours} 
& \textbf{73.5 [62.0, 82.6]} & --
& \textbf{85.2 [67.3, 98.1]} & --
& \textbf{79.6 [67.1, 88.2]} & --
& \textbf{84.7 [70.2, 95.6]} & -- \\
\bottomrule
\end{tabular}
}
\end{table*}

\begin{table*}[h]
\centering
\caption{\textbf{Crayon drawing consistency statistics.} $p$ denotes the raw test $p$-value comparing each method against Ours.}
\label{tab:app_crayon_stats}
\resizebox{0.7\textwidth}{!}{
\begin{tabular}{lcccccc}
\toprule
\multirow{2}{*}{Method}
& \multicolumn{3}{c}{Seen Crayons}
& \multicolumn{3}{c}{Unseen Crayons} \\
\cmidrule(lr){2-4} \cmidrule(lr){5-7}
& Score & 95\% CI & $p$
& Score & 95\% CI & $p$ \\
\midrule
GenDP & 0.81 & [0.71, 0.89] & 0.161
& 0.60 & [0.43, 0.77] & 0.126 \\
Raw Tactile & 0.76 & [0.67, 0.86] & 0.066
& 0.61 & [0.48, 0.74] & 0.016 \\
Sim-Only CF & 0.80 & [0.72, 0.88] & 0.085
& 0.76 & [0.65, 0.85] & 0.313 \\
Real-Only CF & 0.68 & [0.49, 0.85] & 0.087
& 0.74 & [0.56, 0.89] & 0.604 \\
No Force & 0.77 & [0.59, 0.92] & 0.297
& 0.76 & [0.58, 0.91] & 0.724 \\
\textbf{Ours} & \textbf{0.86} & \textbf{[0.77, 0.94]} & --
& \textbf{0.78} & \textbf{[0.66, 0.89]} & -- \\
\bottomrule
\end{tabular}
}
\end{table*}

\begin{table*}[h]
\centering
\caption{\textbf{Peeler policy evaluation statistics.} Contact and cut-in values are success percentages with Wilson score intervals. Peel length is reported in centimeters. $p$ denotes the raw test $p$-value comparing each method against Ours.}
\label{tab:app_peeler_stats}
\resizebox{\textwidth}{!}{
\begin{tabular}{llcccccc}
\toprule
Split & Method 
& Contact [95\% CI] & $p$ 
& Cut-in [95\% CI] & $p$
& Peel Length [95\% CI] & $p$ \\
\midrule
\multirow{6}{*}{Seen}
& GenDP & 45.0 [25.8, 65.8] & 0.024 & 30.0 [14.6, 51.9] & 0.013 & 1.50 [0.33, 3.13] & 0.023 \\
& Raw Tactile & 45.0 [25.8, 65.8] & 0.024 & 20.0 [8.1, 41.6] & 0.002 & 1.05 [0.03, 2.53] & 0.003 \\
& Sim-Only CF & 60.0 [38.7, 78.1] & 0.150 & 30.0 [14.6, 51.9] & 0.013 & 2.00 [0.60, 3.65] & 0.024 \\
& Real-Only CF & 60.0 [38.7, 78.1] & 0.150 & 15.0 [5.2, 36.0] & $<$0.001 & 0.93 [0.00, 2.38] & 0.003 \\
& No Force & 65.0 [43.3, 81.9] & 0.240 & 30.0 [14.6, 51.9] & 0.013 & 1.95 [0.43, 3.90] & 0.032 \\
& \textbf{Ours} & \textbf{80.0 [58.4, 91.9]} & -- & \textbf{70.0 [48.1, 85.5]} & -- & \textbf{4.73 [2.43, 7.10]} & -- \\
\midrule
\multirow{6}{*}{Unseen}
& GenDP & 50.0 [33.2, 66.9] & 0.001 & 33.3 [19.2, 51.2] & 0.002 & 1.12 [0.40, 2.02] & 0.001 \\
& Raw Tactile & 40.0 [24.6, 57.7] & $<$0.001 & 30.0 [16.7, 47.9] & 0.001 & 0.85 [0.20, 1.78] & $<$0.001 \\
& Sim-Only CF & 50.0 [33.2, 66.9] & 0.001 & 46.7 [30.2, 63.9] & 0.032 & 3.05 [1.67, 4.55] & 0.080 \\
& Real-Only CF & 56.7 [39.2, 72.6] & 0.004 & 26.7 [14.2, 44.5] & $<$0.001 & 1.78 [0.70, 3.07] & 0.003 \\
& No Force & 40.0 [24.6, 57.7] & $<$0.001 & 13.3 [5.3, 29.7] & $<$0.001 & 1.08 [0.23, 2.18] & $<$0.001 \\
& \textbf{Ours} & \textbf{90.0 [74.4, 96.5]} & -- & \textbf{73.3 [55.6, 85.8]} & -- & \textbf{4.52 [3.00, 6.05]} & -- \\
\bottomrule
\end{tabular}
}
\end{table*}

\FloatBarrier

\end{document}